\documentclass{article}
\pdfoutput=1
\PassOptionsToPackage{numbers, compress}{natbib}
\usepackage[final]{neurips_2020}
\usepackage[utf8]{inputenc}
\usepackage[T1]{fontenc}
\usepackage{url}
\usepackage{booktabs}
\usepackage{amsfonts}
\usepackage{nicefrac}
\usepackage{microtype}
\usepackage{amsmath,amssymb}
\usepackage{graphicx}
\usepackage{hyperref}
\usepackage[noabbrev]{cleveref}
\usepackage{xcolor}
\usepackage{xspace}
\usepackage[font=small,labelfont=bf]{caption}

\usepackage{tikz}
\usetikzlibrary{tikzmark}

\newcommand{\ballsinbowl}{\mbox{\textsc{BallsInBowl}}\xspace}
\newcommand{\cars}{\mbox{\textsc{RealTraffic}}\xspace}
\newcommand{\genesis}{\mbox{GENESIS}\xspace}
\newcommand{\shapestacks}{\mbox{ShapeStacks}\xspace}
\newcommand{\blockgan}{\mbox{BlockGAN}\xspace}
\newcommand{\ocf}{\mbox{OCF}\xspace}
\newcommand{\blockgantwod}{\mbox{BlockGAN2D}\xspace}
\newcommand{\clevr}{\mbox{CLEVR}\xspace}

\newcommand{\eg}{\textit{e.g.},\xspace}

\makeatletter
\renewcommand{\paragraph}{%
  \@startsection{paragraph}{4}%
  {\z@}{0em}{-1em}%
  {\normalfont\normalsize\bfseries}%
}
\makeatother

\title{RELATE: Physically Plausible Multi-Object Scene Synthesis Using Structured Latent Spaces}
\newcommand*{\affmark}[1][*]{\textsuperscript{#1}}

\author{S\'ebastien Ehrhardt \affmark[1]\thanks{indicates equal contribution} \hspace{2em} Oliver Groth\affmark[1]\footnotemark[1] \hspace{2em}  \'Aron Monszpart\affmark[2,]\affmark[3]\  \hspace{2em} 
Martin Engelcke\affmark[1] \hspace{2em} 
\\
\textbf{Ingmar Posner\affmark[1] \hspace{2em}  Niloy J. Mitra\affmark[2,]\affmark[4] \hspace{2em} Andrea Vedaldi\affmark[1]}\\
\affmark[1]Department of Engineering Science, University of Oxford\\
\affmark[2]Department of Computer Science, University College London\\
\affmark[3]Niantic, \affmark[4] Adobe Research \\
\texttt{\{hyenal,ogroth\}@robots.ox.ac.uk} \\
}

\begin{document}
\maketitle
\begin{abstract}
We present RELATE, a model that learns to generate physically plausible scenes and videos of multiple interacting objects.
Similar to other generative approaches, RELATE is trained end-to-end on raw, unlabeled data.
RELATE combines an object-centric GAN formulation with a model that explicitly accounts for correlations between individual objects.
This allows the model to generate realistic scenes and videos from a physically-interpretable parameterization.
Furthermore, we show that modeling the object correlation is \emph{necessary} to learn to disentangle object positions and identity.
We find that RELATE is also amenable to physically realistic scene editing and that it significantly outperforms prior art in object-centric \emph{scene} generation in both synthetic (\clevr, \shapestacks) and real-world data (cars).
In addition, in contrast to \emph{state-of-the-art} methods in object-centric generative modeling, RELATE also extends naturally to dynamic scenes and generates \emph{videos} of high visual fidelity. Source code, datasets and more results are available at \url{http://geometry.cs.ucl.ac.uk/projects/2020/relate/}.
\end{abstract}

\section{Introduction}\label{s:intro}

We consider the problem of learning to generate plausible images of scenes starting from parameters that are physically interpretable.
Furthermore, we wish to learn such a capability from raw images alone, without any manual or external supervision.
Image generation is often approached via Generative Adversarial Networks (GAN)~\cite{goodfellow14generative}. These models learn to map noise vectors, used as a source of randomness, to image samples.
While the resulting images are  realistic, the random vectors that parameterize them are not interpretable.
To address this issue, authors have recently proposed to \emph{structure} the latent space of deep generative models, giving it a partial physical interpretability~\cite{nguyen2019hologan,nguyen2020blockgan,vansteenkiste2019compositionality}.
For example, \mbox{HoloGAN~\cite{nguyen2019hologan}} samples volumes and cameras to generate 2D images of 3D objects, and \mbox{\blockgan~\cite{nguyen2020blockgan}} creates scenes by composing multiple objects.
The resulting GANs have been shown to learn concepts such as viewpoint and object disentangling from raw images.

BlockGAN is of particular interest because, via its relatively strong architectural biases, it provides \emph{interpretable} parameters for the scene, incorporating concepts such as position and orientation.
However, BlockGAN comes with a significant limitation in that it assumes that objects are mutually \emph{independent}.
This approximation is acceptable only when objects interact weakly, but it is badly violated for medium to densely packed scenes, or for scenes such as stacking wooden blocks or cars following a path, where the (object) correlation is strong.

Recent work in object-centric generative modeling has attempted to specifically address this by capturing correlations in latent space (\eg~\cite{engelcke2019genesis,vansteenkiste2019compositionality}).
However as object state information remains  significantly entangled in these models they have, to date, been unable to operate on real-world data.

In this paper, we introduce RELATE, a model which explicitly leverages the strong architectural biases of BlockGAN to effectively model correlations between latent object state variables.
This leads to a powerful model class, which is able to capture complex physical interactions, while still being able to learn from raw visual inputs alone.
Empirically, we show that only when we model such interactions our GAN model correctly disentangles different objects when they exhibit even a moderate amount of correlation (\cref{fig:ablation,fig:ab-pos}).
Without this component, the model may still generate high fidelity images, but it generally fails to establish a physically-plausible association between the parameters and the generated images.
Our results also demonstrate that GANs are surprisingly sensitive to the correlation of objects in natural scenes, and can thus be used to directly learn these \emph{without} resorting to techniques such as variational auto-encoding (VAE~\cite{kingma2014vae}).

We demonstrate the efficacy of RELATE in several scenarios, including balls rolling in bowls of variable shape~\cite{ehrhardt2019taking}, cluttered tabletops~(\clevr~\cite{johnson2017clevr}), block stacking~(\shapestacks~\cite{groth2018shapestacks}), and videos of traffic at busy intersection.
By ablating the interaction module, we show that modeling the spatial correlation between the objects is key.
Furthermore, we compare RELATE to several recent GAN- and VAE-based baselines, including \blockgan~\cite{nguyen2020blockgan}, \genesis~\cite{engelcke2019genesis} and \ocf~\cite{anciukevicius2020ocf}, in terms of \emph{Fr\'{e}chet Inception Distance (FID)}~\cite{heusel2017fid}, and outperform even the best state-of-the-art model by up to 29 points.

Qualitatively, we show that modeling spatial relationships strongly affects scene decomposition and the enforcement of spatial constraints in the generated images.
We also show that the physically interpretable latent space learned by RELATE can be used to edit scenes as well as to generate scenes outside the distribution of the training data (\eg containing more or fewer objects).
Finally, we show that the parameterization can be used to generate long plausible video sequences (as measured according to FVD score~\cite{unterthiner2018fvd}) by simulating their dynamics while preserving their spatial consistency.

\begin{figure}[t]
\centering
\includegraphics[width=\linewidth]{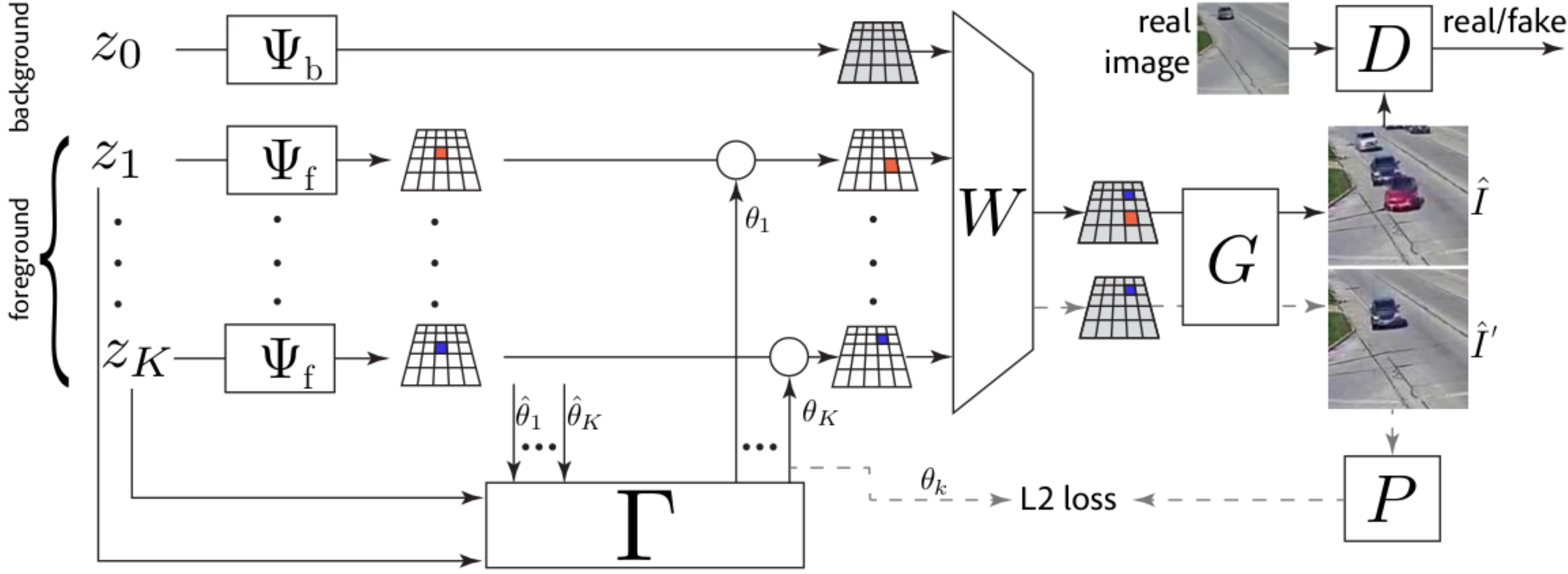}
\caption{
\textbf{Image generation using RELATE.} Individual scene components, such as background and foreground objects are represented by
appearance $z_0$ and pairs of appearance and pose vectors $(z_i, \theta_i), i \in \{1,\dots,K\}$, respectively. The key spatial relationship module $\Gamma$ adjusts the initial independent pose samples $\hat{\theta}_i$ to be physically plausible (\eg non-intersecting) to produce ${\theta}_i$. The structured scene tensor  $W$ is finally transformed by the the generator network $G$ to produce an image $\hat{I}$. RELATE is trained end-to-end in a GAN setup ($D$ denotes the discriminator) on real unlabelled images.
}
\label{fig:architecture-details}
\end{figure}

\section{Related Work}

\paragraph{Interpretable Object-Centric Visual Models.}
Inspired by the \emph{analysis-by-synthesis} approach for visual perception discussed in cognitive science~\citep{yildirim2020objectness}, recent work~\citep{burgess2019monet,engelcke2019genesis,greff2019iodine,vansteenkiste2019compositionality} propose structured latent space models to explain and synthesize images as sets of constituent components which are individually represented using VAEs~\citep{kingma2014vae} or GANs~\citep{goodfellow14generative}.
Other approaches favor explicit symbolic representations over distributed ones when parsing an image~\citep{santoro2017simple, wu2017derendering} or propose probabilistic programming languages to formalize image generation~\citep{kulkarni2015picture}.
In both cases, object-centric modeling allows decomposition of images into components and also enables targeted image modification via interpolation in symbol or latent space, \eg altering position or color of an object parsed by the model.
Interpretable and controllable factors of image generation are desirable properties for neural rendering models and have been investigated in recent image generation models, \eg~\cite{anciukevicius2020ocf,liao2019controllablegan,nguyen2019hologan,nguyen2020blockgan}.
However, despite the modeling effort put into the object representations, inter-object interactions are typically only modeled in a less explicit way, \eg via image layers~\citep{vansteenkiste2019compositionality,wu2017derendering}, depth ordering variables~\citep{anciukevicius2020ocf} or an autoregressive \emph{scene prior}~\citep{engelcke2019genesis}.
Our work adds to this line of work by proposing a spatial correlation network which facilitates disentanglement of learned object representations and can be trained from raw observations.

\paragraph{Neural Physics Approximation.}
Harnessing the power of deep learning to approximate physical processes is an emerging trend in the machine learning community.
Especially the approximation of rigid body dynamics with neural networks already boasts a large body of literature, \eg~\cite{battaglia2016interaction,chang2017compositional,fragkiadaki2016billard,kossen2019stove,vansteenkiste2018relationalnem,watters2017visualinteraction}.
Such learned approximations of object interactions have been successfully employed in object manipulation~\citep{janner2019o2p2} and tracking~\citep{fuchs2019mohart,kosiorek2018sqair}.
However, most entries in this line of work are only applied to visual toy domains or rely on segmentation masks or bounding boxes to initialize their object representations before the networks approximate the object dynamics.
While we leverage ideas from neural dynamics modeling, we go beyond the established scope of visual toy domains such as colored point masses or moving MNIST digits~\citep{lecun1998mnist} and learn directly from rich visual data such as simulated object stacks and real traffic videos without further annotation.

\paragraph{Extraction and Generation of Video Dynamics.}
Videos are a natural choice of data source to learn about the physics of rigid bodies.
Physical information extracted from videos can either be explicit such as estimates of velocity or friction~\citep{ye2018interpretable} or implicitly represented in the latent space, \eg sub-spaces corresponding to pose variation~\citep{denton2017disentangledvideos}.
More recently, object-centric approaches have also been leveraged to acquire better video representations for future frame prediction~\citep{ye2019cvp} or model-based reinforcement learning~\citep{veerapaneni2019op3}.
Several studies have also attempted to learn entire video distributions as spatio-temporal tensors in GAN frameworks~\citep{kalchbrenner2017vpn,saito2017tgan,vondrick2016generating,xue2016visualdynamics} yielding impressive first results for full video generation in artificial and real domains.
In contrast to prior art, our model departs from a monolithic spatio-temporal tensor representation over an entire video.
Instead we cast the video learning and generation process as temporal extension of the object-centric representation of a single frame, lowering the computational burden while still faithfully representing long-range dynamics.

\section{Method}\label{s:method}

RELATE (\cref{fig:architecture-details}) consists of two main components: An interaction module, which computes physically plausible inter-object and object-background relationships, and a scene composition and rendering module, which features an interpretable parameter space factored into appearance and position vectors.
The details are given next.

\subsection{Physically-interpretable scene composition and rendering}

RELATE considers scenes containing up to $K$ distinct objects.
The model starts by sampling \emph{appearance parameters} $z_1,\dots,z_K \sim \mathcal{U}([-1,1]^{N_f})$ for each individual foreground object as well as a parameter $z_0 \sim \mathcal{U}([-1,1]^{N_b})$ for the background.
These parameters are small noise vectors, similar to the ones typically used in generative networks.
Different from the object poses below, they are sampled independently, thus assuming that the appearance of different objects is independent.

For rendering an image, the appearance parameter $z_k$ is first mapped to a tensor $\Psi_k \in \mathbb{R}^{H\times H\times C}$. This is done via two separate learned decoder networks, one for the background $\Psi_0 = \Psi_b(z_0)$ and one for the foreground objects $\Psi_k = \Psi_f(z_k)$.
Here $H$ is the horizontal and vertical spatial resolution of the representation (see \cref{tab:hyperparams} in supplementary) and $C$ is the number of feature channels (see \cref{tab:psib,tab:psif}).
Since we assume that individual objects are much smaller than the overall scene, we restrict $\Psi_k$, $k \geq 1$ to be non-zero only in a fixed smaller $H' < H$ window in the center of the tensor.

Each foreground object also has a corresponding \emph{pose parameter} $\theta_k$, which is geometrically interpretable.
For simplicity, we assume $\theta_k \in \mathbb{R}^2$ to be a 2D translation, acting on the tensor $\Psi_k$ via bilinear resampling:
$$
\hat\Psi_k = \theta_k \cdot \Psi_k
~~~\text{such that}~~~
[\hat \Psi_k]_{u}  = [\Psi_k]_{u + \theta_k}
$$
where $u\in\mathbb{R}^2$ is a spatial index and $[\cdot]_u$ means accessing the column of the tensor in bracket at spatial location $u$ (using padding and bilinear interpolation if $u$ does not have integer coordinates).
However, $\theta_k$ can easily be extended to represent full 3D transformations as previously shown in \blockgan~\cite{nguyen2020blockgan}.

Foreground and background objects are composed into an overall scene tensor $W \in \mathbb{R}^{H \times H \times C}$ via element-wise max- (or sum-) pooling as
$
  W_u =  \max_{k=0,\dots,K} [\hat \Psi_k]_u.
$
In this manner, the scene tensor is a function
$
 W(\Theta,Z)
$
of the pose parameters $\Theta:=(\theta_1,\dots,\theta_K)$ and the appearance parameters $Z:=(z_0,z_1,\dots,z_K)$.
Finally, a decoder network $\hat I = G(W)$ renders the composed scene as an image (see \cref{tab:generator}).

\paragraph{Discussion.}
This model is `physically interpretable' in the sense that it captures
(1) the identities of $K$ distinct objects and
(2) their pose parameters as translation vectors.
This should be contrasted to traditional GAN models, where the code space is given as an uninterpretable, monolithic noise vector $z$.
Despite the structure given to the code space, there is no guarantee that the model will actually learn to map it to the corresponding structure in the example images.
However, we found empirically that this is the case as long as the correlations between the different objects are also captured.

\subsection{Modeling correlations in scene composition}
RELATE departs significantly from prior art such as \blockgan as it does not assume the parameters $\theta_i$ of the different objects to be independent.
In order to model correlation, we propose a two-step procedure, based on a residual sampler.
First, we sample a vector of $K$ i.i.d.~poses $\hat \Theta \sim \mathcal{U}([-H''/2,H''/2]^{2K})$ where $H'' < H$ is smaller than the spatial size $H$ of the tensor encoding.
Then, we pass this vector to a `correction' network $\Gamma$ that remaps the initial configuration to one that accounts for the correlation between object locations and appearances, as well as between objects and the background (coded by the appearance component $z_0$ in $z$):
$
  \Theta := \Gamma(\hat\Theta, Z).
$
In practice, we expect object interactions, as any physical law, to be \emph{symmetric} with respect to the order of the objects.
We obtain this effect by implementing $\Gamma$ as running $K$ copies of the \emph{same} corrective function in parallel:
\begin{equation}\label{eq:theta}
      \theta_k = \hat \theta_k + \zeta(\hat \theta_k, z_k, | z_0, \{z_i, \hat\theta_i \}_{i \geq 1, i\neq k}).
\end{equation}
The function $\zeta$ is implemented in a manner similar to the Neural Physics Engine (NPE)~\cite{chang2017compositional}:
\begin{equation}
\zeta(\hat \theta_k, z_k, | z_0, \{z_i, \hat\theta_i \}_{i \geq 1, i\neq k})
= f(\hat \theta_k, z_k, z_0, h_k^s),
~~~~
h_k^s = \sum_{q\neq k} g(\hat \theta_k, z_k, \hat \theta_q, z_q),
\end{equation}
where $f$ and $g$ are Multi Layer Perceptrons (MLPs) (tables \ref{tab:f}, \ref{tab:g}) operating on stacked vector inputs and $h^s$ is an embedding capturing the interactions between the $K$ objects.
Besides symmetry, an advantage of this scheme is that it can take an arbitrary number of objects $K$ due to the sum-pooling operator used to capture the interactions.
In this manner, the sampler $\Gamma$ is automatically defined for any value of $K$. For each scene, K is sampled uniformly from a fixed interval $[K_\text{min}, K_\text{max}]$.
Furthermore, sampling independent quantities followed by a correction has the benefit of injecting some variance on the objects positions at the early stage of training, which helps to avoid converging to trivial/bad solutions.

\paragraph{Ordered scenes.}

An advantage of RELATE is that it can be easily modified to take advantage of additional structure in the scene.
For scenes where objects have natural order, such as stacks of blocks, we experiment with conditioning pose $\theta_i$ on the preceding pose $\theta_{i-1}$, using a Markovian process.
This is done by first sampling $\hat \theta_1  \sim \mathcal{U}([-H''/2,H''/2])$, and then applying a correction to account for the background $z_0$ as before, finally sampling the other objects in sequence:
\begin{equation}
\theta_1 = \hat \theta_1 + f_0(\hat \theta_1, z_1, z_0),
~~~~~~~
\forall k > 1 :
~~~
\theta_k = \theta_{k-1} + f_1(\theta_{k-1}, z_{k-1}, z_0),
\end{equation}
where $f_0, f_1$ are implemented as MLPs as before (tables \ref{tab:f0}, \ref{tab:f1}).
Note that this can be interpreted as a special case of the model above in the sense that we can write
$
  \Theta := \Gamma(\hat\Theta, Z),
$
provided that $\hat\theta_k = 0$ for $k \geq 2$.

\paragraph{Modeling dynamics.}
RELATE can also be immediately extended to make dynamic predictions.
For this, we sample the initial positions $\theta_k(0)$ as before and then update them incrementally as
$
\theta_k(t+1) = \theta_k(t) + v_k(t+1),
$
where $v_k(t)$ is the object velocity.
In order to obtain the latter, we let
$
V_k(t)= [v_k(t-i)]_{i=2,1,0}
$
denote the last three velocities of the $k$-th object.
The initial value $V_k(0) =e_v(z_k, z_0, \theta_k(0))$ is initialized as a function of the appearance parameters and initial positions (\cref{tab:e}); and we use the NPE style update equations~\cite{chang2017compositional}, where $e_v$, $f_v$ and $g_v$ are MLPs,
\begin{equation}\label{eq:theta_v}
v_k(t+1)
= f_v(\theta_k(t), z_k, V_k(t),z_0, h_k^d(t)),
~~~
h_k^d(t) = \sum_{q\neq k} g_v(\theta_k(t), z_k,  V_k(t), \theta_q(t), z_q,  V_q(t)).
\end{equation}

\subsection{Learning objective}
Training our model makes use of a training set $I_i$, $i=1,\dots,N$ of $N$ images of scenes containing different object configurations.
No other supervision is required.
Our learning objective is a sum of \emph{two high fidelity losses} and \emph{a structural loss} which we describe below.

For high fidelity, images $\hat I$ generated by the model above are contrasted to real images $I$ from the training set using the standard GAN discriminator $\mathcal{L}_{\text{GAN}}(\hat I, I)$ and style $\mathcal{L}_{\text{style}}(\hat I, I)$ losses from~\cite{nguyen2020blockgan} (see \cref{sec:losses}).

In addition, we introduce a regularizer to encourage the model to learn a non-trivial relationship between object positions and generated images.
For this, we train a position regressor network $P$ that, given a generated image $\hat I$, predicts the location of the objects in it.
In practice, we simplify this task and generate an image $\hat I'$ by retaining only object $k$ of the $K$ objects at random and minimizing
$\| \check{\theta}_k - P(G(W(z_0, z_k, \theta_k))) \|^2_2$.
Here the symbol $\check \cdot$ means that gradients are not back-propagated through $\theta_k$:
this is to avoid mode collapse of the position at zero. $P$ shares most of its weights with the discriminator network (see \cref{tab:decoder}).

In the case of dynamic prediction, the discriminator takes as input the sequence of images concatenated along the RGB dimension and is tasked to discriminate between fake and real sequences. Similar to a static model we also have a position regressor which is tasked to predict the position of an object rendered at random with zero velocity.

\section{Experiments}\label{s:experiments}

\begin{figure}[t] 
\begin{minipage}{0.65\linewidth}
\centering
\includegraphics[width=\textwidth]{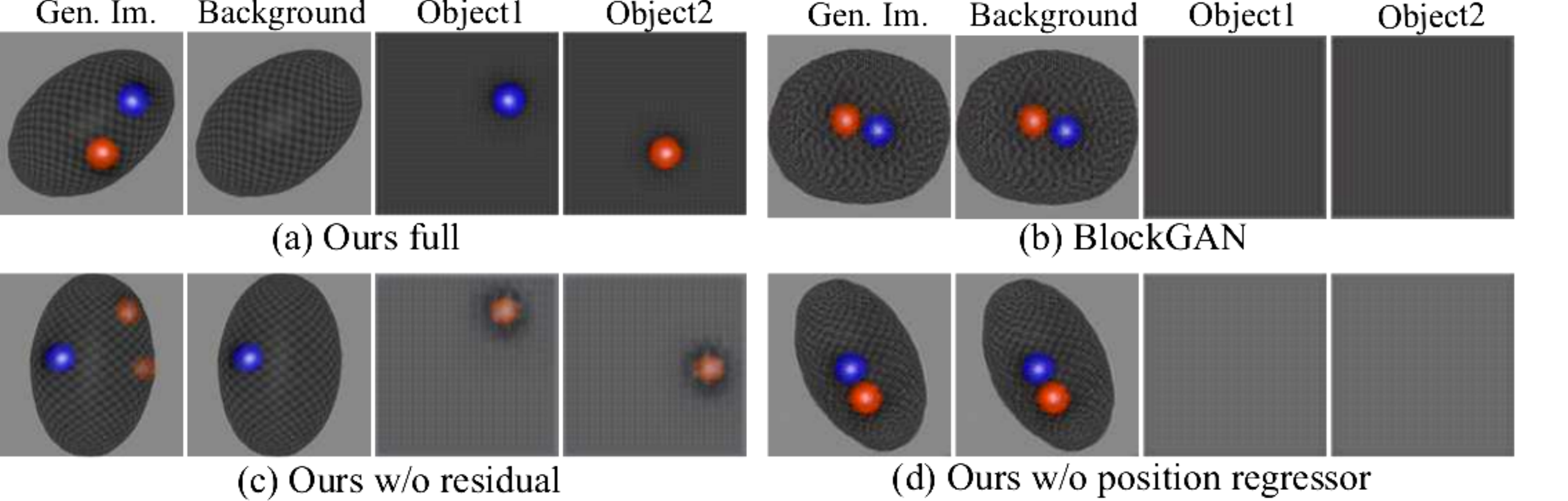}
\captionof{figure}{\textbf{Ablation Study.} For every case we render every component of our method independently. We show that only our full model is able to correctly disentangle individual components of the scene.}\label{fig:ablation}
\end{minipage} 
\begin{minipage}{0.33\linewidth}
\centering
\captionof{table}{\textbf{Ablation study.} FID score (lower is better) on \ballsinbowl. Ours (full) reaches the highest fidelity by a large margin.}\label{tab:ablation}
\scriptsize
\begin{tabular}{lr}
\toprule
BlockGAN~2D\cite{nguyen2020blockgan} & 152.3\\
\midrule
Ours w/o residual                    & 133.9  \\
\midrule
Ours w/o pos.~reg.                   & 154.8 \\
\midrule
Ours (full)                          & \textbf{81.9} \\
\bottomrule  
\end{tabular}
\end{minipage}
\end{figure}
\begin{figure}[t]
\centering
\includegraphics[width=\textwidth]{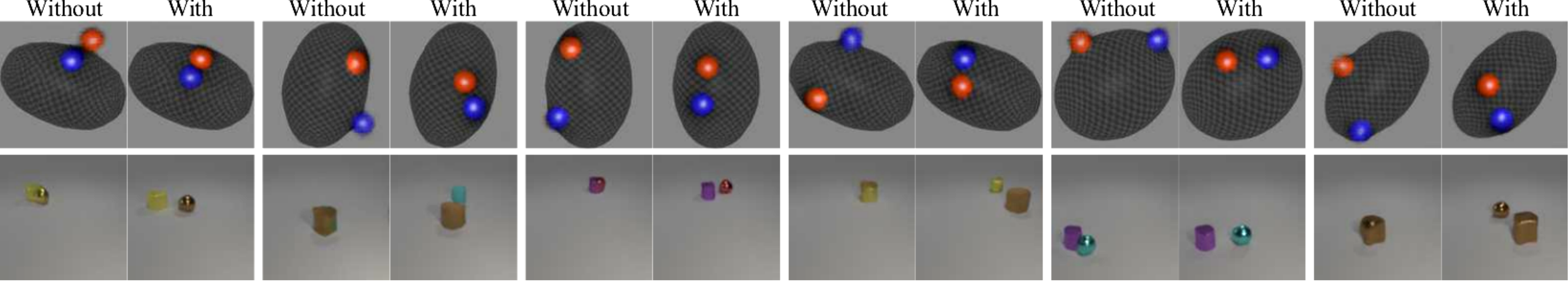}
\caption{\textbf{Effect of the interaction module $\Gamma$.}
We show pairs of images without and with the correction function $\Gamma$ is applied.
For \ballsinbowl the correction moves the balls within the bowl, and for \clevr it pushes apart intersecting objects.}\label{fig:ab-pos}
\end{figure}

\begin{table}[t]
\centering
\caption{{\bf Comparison to state-of-the-art methods.} FID score (lower is better) for various datasets. We consistently outperfom prior art in object centric scene generation. `Ordered' refers to the variant discussed in sec. 3.2. \dag are standard GANs which FID score are evaluated on $64\times64$ images for `General' variant (see \cref{sec:app:impldet}).\ddag is the 2D variant of BlockGAN~\citep{nguyen2020blockgan} which sometimes fails to be object-centric (see \cref{sec:dec}).}\label{t:sota}
\small
\begin{tabular}{lccccc}
\toprule
                                         & CLEVR-5       & CLEVR-5vbg    & \clevr        & \shapestacks   & \cars\\
RELATE variant                           & General       & General       & General       & Ordered        & General \\
\midrule
DCGAN\dag & 264.8 & 361.8 & 247.8 & 197.6 & 47.6 \\

DRAGAN\dag & 80.8 & 84.4 & 108.0  & \textbf{57.2} & \textbf{38.8}\\
\midrule
OCF~\cite{anciukevicius2020ocf}          & N/A           & 83.1          & N/A           & N/A            & N/A\\
\genesis~\cite{engelcke2019genesis}      & 211.7         & 169.4         & 151.3         & 233.0          & 167.1 \\
\blockgantwod \ddag~\cite{nguyen2020blockgan} & 63.0          & 53.3         & 78.1          & 99.3          & 57.9\\
\midrule
Ours                                     & \textbf{58.4} & \textbf{36.4} & \textbf{62.9} & \textbf{95.8} & \textbf{42.0}\\
\midrule
Ours + scale (see \cref{ssec:scale})  & \textbf{37.4} & \textbf{35.9} & \textbf{44.9} & \textbf{79.1} & 46.8\\
\bottomrule
\end{tabular}
\end{table}

\begin{figure}[t]
\centering
\includegraphics[width=\linewidth]{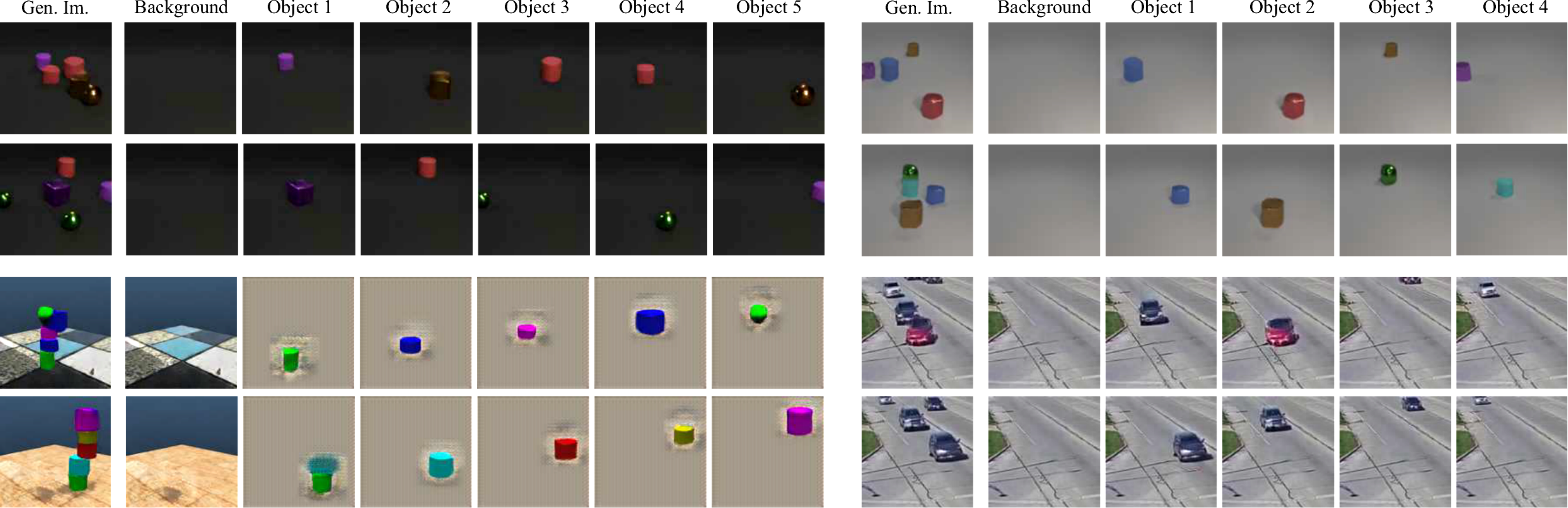}
\caption{\textbf{Component-wise scene generation.} From a generated image (left) RELATE can render each component individually for each dataset. For \clevr and \cars objects are rendered after being composed with the background (cf.~\cref{sec:interpretability}). Top left picture has increased contrast for easier visualization.}
\label{fig:scene-dec}
\end{figure}
\begin{figure}[t]
\centering
\includegraphics[width=\linewidth]{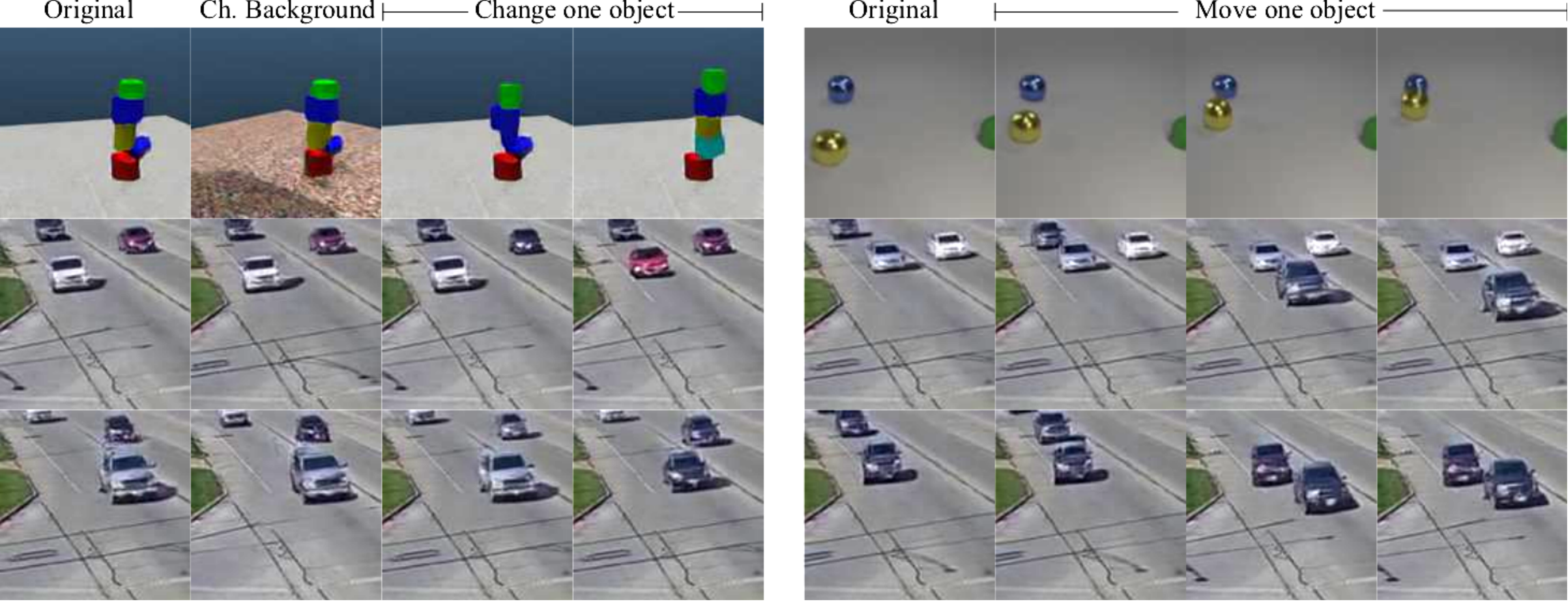}
\caption{\textbf{Image editing.}
Left: We demonstrate the capacity of RELATE to change the background and  the appearance of individual objects.
Right: RELATE is also able to modify the position of a single object.}\label{fig:edit}
\end{figure}
\begin{figure}[t]
\centering
\includegraphics[width=\linewidth]{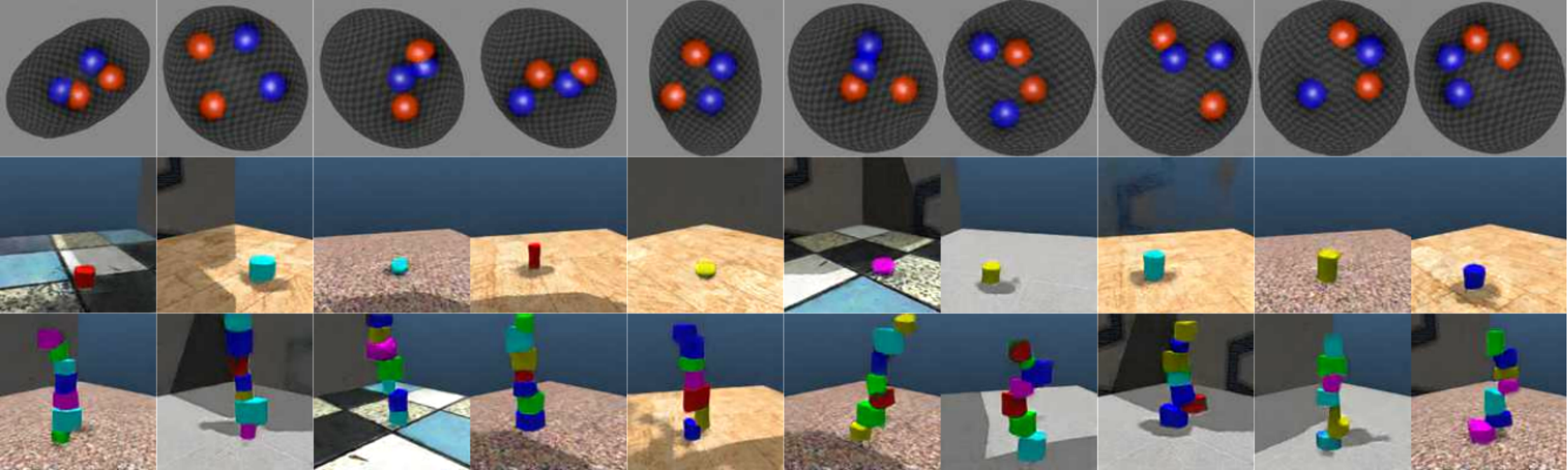}
\caption{\textbf{Out-of distribution generation.} RELATE can generate images outside the training distribution.
The first row shows generating a bowl with four balls, whereas the training set only features exactly two.
The last two rows depict towers of one and seven objects, whereas the training images only had stacks of height two to five.}\label{fig:out}
\end{figure}

\paragraph{Implementation details.}

We learn mappings $\Psi_b$ and $\Psi_f$ using the same Adaptive Instance Normalization (AdaIN)~\cite{huang2017arbitrary} architecture.
The spatial size of their output tensors is set to $H=16$ and the final output image to $128\times 128$ (which is reduced when needed for fair comparison to other methods).
We use the Adam~\cite{kingma15adam} optimizer for learning and train for a fixed number of epochs and always select the last model snapshot.
We consider two types of baselines: standard generative models such as DCGAN\cite{radford2015unsupervised} and DRAGAN\cite{kodali2017convergence}, and object-centric generative baselines such as \genesis~\cite{engelcke2019genesis} and \mbox{OCF~\cite{anciukevicius2020ocf}}, quoting results from the original papers whenever possible. In addition we also add \blockgantwod as an ablation of our method.
\paragraph{Datasets.}

We conduct experiments on four different datasets.
First, we consider a relatively simple dataset, \ballsinbowl~\cite{ehrhardt2019taking}, for assessing the model features and ablations.
It consists of videos of two distinctly colored balls rolling in an elliptical bowl of variable orientation and eccentricity.
Interactions amount to object collisions and the fact that they must roll within the bowl.
To this, we add two popular synthetic datasets
\mbox{\clevr~\cite{johnson2017clevr}} (cluttered tabletops) and \mbox{\shapestacks~\cite{groth2018shapestacks}} (block stacking).
Finally, we have collected a new dataset \cars containing five hours of footage of a busy street intersection, divided into fragments containing from one to six cars.
Especially the last dataset contains many interactions between the individual cars as they adapt their speed to the surrounding traffic which happens frequently when the light changes and cars either slow down because of a queue on red or accelerate when the lights change to green again.
Further details about training, evaluation and datasets can be found in the appendix, sections \ref{sec:app:impldet}, \ref{sec:baselines} and \ref{sec:dataset}.

\subsection{Generating static scenes}\label{s:static}

\paragraph{Ablation study.}

We start experimenting with the comparatively simple \ballsinbowl dataset to conduct basic ablations.
The first ablation removes the spatial correlation module $\Gamma$ and the position regression loss, therefore reducing RELATE to a 2D version of BlockGAN\@.
We also consider \textit{`w/o residual'}, where the addition $\hat \theta_k$ in \cref{eq:theta} is removed, and \textit{`w/o pos.~loss'}, where the position regression loss regularizer is removed.
\Cref{tab:ablation} shows that each component of RELATE yields an improvement in terms of FID scores on this dataset supporting our spatial modeling decisions.
Furthermore, in~\cref{fig:ablation} we show qualitatively that only RELATE (a) is able to correctly disentangle the underlying scene factors.
We do this by generating the same image while retaining a single factor, which correctly isolates the background, and, in turn, both individual objects.
BlockGAN 2D (b) and \textit{`ours w/o pos.~reg.'} (d) fail to disentangle the factors entirely, mapping everything to the background component.
\textit{`Ours w/o residual'} (c) shows that the model partially fails to disentangle, with the background encoding some but not all the objects.
Finally, \cref{fig:ab-pos} visualizes the effect of the interaction module $\Gamma$.
Recall that this is implemented as a `correction' function that accounts for correlation starting from independently-sampled parameters.
For \ballsinbowl, the correction module moves the balls within the bowl, and for the \clevr it pushes objects apart if they intersect.

\paragraph{Quantitative evaluation.}

In~\cref{t:sota}, we compare RELATE to existing scene generators on \shapestacks, \clevr and \cars.
We report performance in terms of FID score~\cite{heusel2017fid} computed between 10,000 images sampled from our model and the respective test sets.
For \clevr, we train RELATE and BlockGAN on a restricted version of the data containing from three to six objects in an image\footnote{Note that \genesis was trained on the full training set featuring three to ten objects.},
and at test time, we require all models to sample images with three to ten objects.
We consistently outperform all prior object-centric methods in all scenes and scenarios according to FID scores.
In particular, on \clevr, RELATE can generate a larger number of objects than seen during training suggesting its improved generalization capabilities which are demonstrated further in~\cref{fig:out}.
Our method also consistently out-performs standard GANs on all CLEVR datasets and is on par with DRAGAN on \cars.
More qualitative results can be found in \cref{sec:qr}.

\subsection{Interpretability of the latent space and scene editing}
\label{sec:interpretability}

As shown in the ablation studies in~\cref{fig:ablation}, RELATE successfully disentangles a scene into independent components - in contrast to \blockgantwod which struggles to separate individual objects from the background.
\Cref{fig:scene-dec} shows that RELATE can \emph{disentangle} also far more complex scenes in \cars, \shapestacks and \clevr.
Note that for \cars and \clevr we render objects composed with the background.
In fact, in these datasets the size and appearance of each object is correlated to their position in the background because of the camera perspective.
In addition to qualitative results, we also compute a disentanglement score in \cref{t:dis} which measures how well our model is disentangling individual components of the scene.
We found that our model manages to consistently separate each individual objects of the scene and outperform \blockgantwod on the most challenging datasets which is in line with the qualitative evidence we observe.
Next, in \cref{fig:edit} we use RELATE to \emph{edit} a generated scene.
For example we can change the position or appearance of individual objects.
Finally, we show that RELATE can generate \emph{out-of-distribution} scenes.
This is achieved in particular by sampling a different number of objects.
In \cref{fig:out}, for instance, RELATE is trained on \shapestacks seeing towers of height two to five.
However, it can render taller towers of up to seven objects, or even just a single object.
Likewise, in \ballsinbowl it can generate bowls with four balls having seen only two during training.
Furthermore, in \shapestacks each tower is composed of blocks of \emph{different} colors, but RELATE can relax this constraint rendering objects with repeated colors.

\subsection{Simulating dynamics}\label{ssec:dyn}

\begin{figure}
\centering
\includegraphics[width=\linewidth]{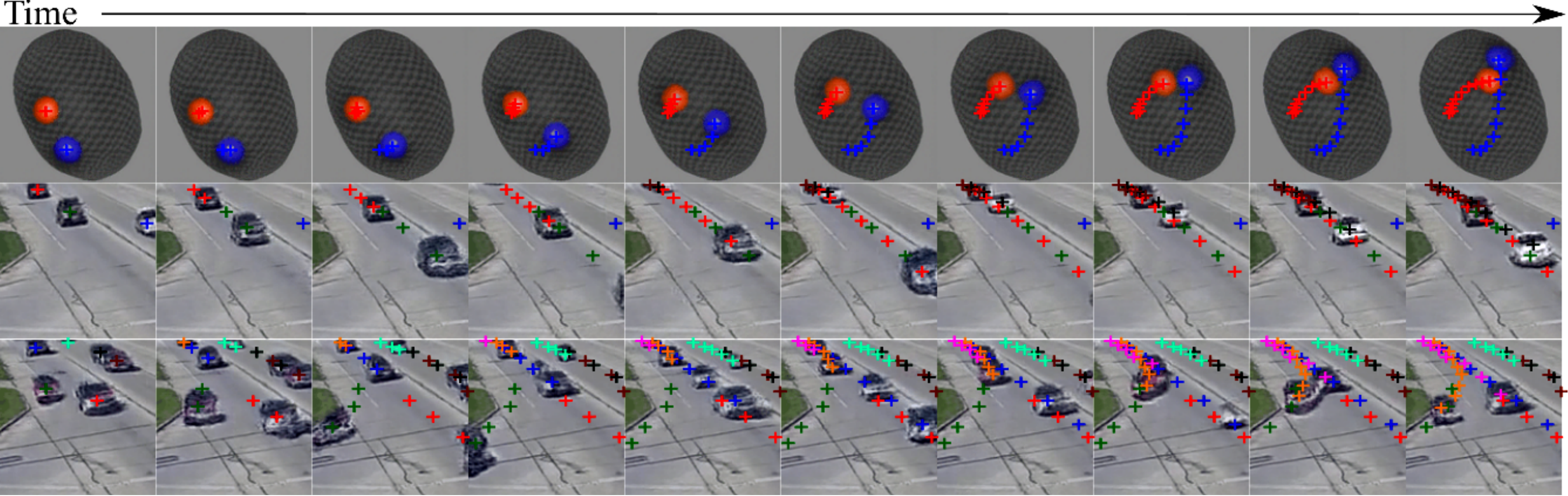}
\caption{\textbf{Video generation.}
We show consecutive video frames generated by RELATE overlayed with crosses representing projections of the model's estimated pose parameters for each object\@.
In \ballsinbowl the interaction with the environment is well captured as the balls stay within the bowl.
In \cars the cars stay in their lane, or can decide to make a right turn (last row).}\label{fig:video}
\end{figure}

We train this model on \ballsinbowl and \cars to predict 15 and 10 consecutive frames respectively.
During generation, we sample videos with a sequence length of 30 frames and measure the faithfulness with respect to the distribution of the test data via the \emph{Fr\'{e}chet Video Distance (FVD)}~\cite{unterthiner2018fvd}.
We achieve FVD scores of $556$ and $2253$ respectively. This is perceptibly better than $920$ and $3370$ for a baseline consisting of time-shuffled sequences from the respective training sets, which feature perfect resolution but poor dynamics.
Qualitatively in \cref{fig:video} we see that the model does understand the motion and captures interaction with the background.
For instance, in \ballsinbowl the balls do have a curved motion because of the shape of the bowl and decrease in speed when reaching the edges of the bowl which are in higher position (see first row of \cref{fig:video}).
In \cars the cars do stay in their respective lane.
Interestingly our model is able to handle different types of motions correctly (see third row \cref{fig:video}) and uses the sample vector to decide whether the cars should go straight or turn.
Finally, we see that we can also generate videos with much more cars than the upper bound (5) with which the system was trained (see last row \cref{fig:video}).

\begin{figure}
\centering
\includegraphics[width=\linewidth]{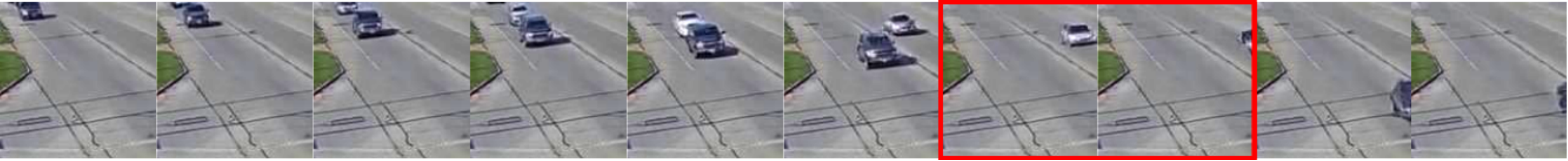}
\caption{\textbf{Failure case.}
Our model struggles to understand big changes of aspect ratio.
For instance, in the static case, when we drag two cars towards the bottom of the image, we can see that the left grey car disappears at some point (highlighted in red) and reappears on the edge of the image.
This explains why the video generation on \cars fails to capture the true data distribution more faithfully resulting in lower FVD score.
This effect can however be mitigated with a scale augmented model (see \cref{ssec:dyn}).}
\label{fig:limitations}
\end{figure}

\paragraph{Limitations}

While the dynamics in video generation look realistic in most cases, \cars also exposed the limitations of our approach.
In this dataset the perspective range of the camera is important.
As a result, cars at the bottom of the image, which appear bigger in the training data, often do not get generated properly in the static case (see highlighted frames in \cref{fig:limitations}).
We hypothesize that this is also the main reason why the cars' appearance (and hence FVD score) deteriorates in the dynamic scenarios: since the style parameters $z_i$ are fixed to preserve identity, it is not possible for the model to account accurately for the appearance change introduced by large changes of perspective over the course of a sequence.
To verify our hypothesis in \cref{ssec:scale} we propose a simple modification of our pipeline that accounts for scale changes in an image. This modification simply allows the network to predict $H'$ for each individual object instead of it being a constant. We found that in general this allows our network to train on datasets with more important range of scales such as CLEVR3~\cite{anciukevicius2020ocf} and renders objects at more different scales for \clevr and \cars (see \cref{fig:relate_scale}). Besides in \cref{t:sota} we show that this modification almost always results in higher FID scores from our main model as well as a significant boost of 397 points in FVD score for \cars (1855 vs 2253).

\section{Conclusion}

We have introduced RELATE, a GAN-based model for object-centric scene synthesis featuring a physically interpretable parameter space and explicit modeling of spatial correlations.
Our experimental results suggest that spatial correlation modeling plays a pivotal role in disentangling the constituent components of a visual scene.
Once trained, RELATE's interpretable latent space can be leveraged for targeted scene editing such as altering object positions and appearances, replacing the background or even inserting novel objects.
We demonstrate our model's effectiveness by presenting \emph{state-of-the-art} scene generation results across a variety of simulated and real datasets.
Lastly, we show how our model naturally extends to the generation of dynamic scenes being able to generate entire videos from scratch.
A main limitation of our current model is its restriction to planar motions which prevents it from representing arbitrary 3D motions featuring angular rotation more faithfully, most notably highlighted by the experiments for video generation.
This effect can however be partially mitigated by a scale-augmented model which we introduce in the appendix.
We believe that our work can contribute to future research in object-centric scene representation by providing a scalable, spatio-temporal modeling approach which is conveniently trainable on unlabeled data.
\section*{Broader Impact}

Our method advances the ability of computers to learn to understand  environments in images in an object-centric way.
It also enhances the capabilities of generative models to generate realistic images of ``invented'' environment configurations.

Overall, we believe our research to be at low to no risk of direct misuse.
At present, our generation results are insufficient to fool a human observer.
However, it has to be noted that the sampling process is, as in many other deep generative models, capable of revealing patterns observed in the training data, \eg specific textures or object geometries.
Such data privacy concerns are not applicable in the street traffic data used in our research, since the resolution of the videos is far too low to identify individual drivers or recognize cars' license plates.
However, `training data leakage' should be taken into consideration when the model is trained on more sensitive datasets.

In a positive prospect, we believe that our model contributes to further the development of less opaque machine learning models.
The explicit object-centric modelling of image components and their geometric relationships is in many of its aspects intelligble to a human user.
This facilitates debugging and interpreting the model's behaviour and can help to establish trust towards the model when employed in larger application pipelines.

However, the key value of our paper is in the methodological advances.
It is conceivable that, like any advance in machine learning, our contributions could ultimately lead to methods that in turn can and are misused.
However, there is nothing to indicate that our contributions facilitate misuse in any direct way; in particular, they seem extremely unlikely to be misused directly.

\section*{Acknowledgments} 
This work is supported by the European Research Council under grants ERC 638009-IDIU, ERC 677195-IDIU, and ERC 335373.
The authors acknowledge the use of Hartree Centre resources in this work. The STFC Hartree Centre is a research collaboratory in association with IBM providing High Performance Computing platforms funded by the UK's investment in e-Infrastructure.
The authors also acknowledge the use of the University of Oxford Advanced Research Computing (ARC) facility in carrying out this work (\texttt{http://dx.doi.org/10.5281/zenodo.22558}).
Special thanks goes to Olivia Wiles for providing feedback on the paper draft, Thu Nguyen-Phuoc for providing the implementation of BlockGAN and Titas Anciukevi\v{c}ius for providing the generation code for CLEVR variants.
We finally would like to thank our reviewers for their diligent and valuable feedback on the initial submission of this manuscript.

\bibliographystyle{plain}
\bibliography{oxgan}
\clearpage
\setcounter{section}{0}
\renewcommand\thesection{A\arabic{section}}
\setcounter{table}{0}
\renewcommand{\thetable}{A\arabic{table}}

\setcounter{figure}{0}
\renewcommand{\thefigure}{A\arabic{figure}}

\setcounter{page}{1}

\begin{center}
     \Large\bf
     Supplementary ~material ~for ``RELATE\@: Physically plausible Multi-Object Scene Synthesis Using Structured Latent Spaces''
\end{center}

In this supplementary material we provide further details about RELATE.
The appendix is organised as follows:
First, we provide an additional object decomposition score and a description of the scale experiment in \cref{sec:dec}.
Further discussions about the method are presented in \cref{sec:lim_app}.
Details on the exact loss functions used can be found in \cref{sec:losses}.
In \cref{sec:app:impldet} we elaborate on the model's implementation details.
\Cref{sec:baselines} is dedicated to explaining the details of the baselines and their respective training protocols.
\Cref{sec:dataset} contains a thorough explanation of every dataset and the data collection procedure where applicable.
Finally, we provide more qualitative results in \cref{sec:qr}.

\section{Additional experiments}\label{sec:dec}

\subsection{Disentanglement study}
\begin{table}[htb]
\centering
\caption{{\bf Disentanglement score.} For each dataset of \cref{t:sota} we report respectively the distance and correlation score described in \cref{sec:dec}. Our model outperforms BlockGAN2D in the most complex scenario: \shapestacks and \ballsinbowl. Both model reach similar scores for the other scenes.}\label{t:dis}
\small
\begin{tabular}{p{1.3cm}cccccc}
\toprule
                                         & CLEVR-5       & CLEVR-5vbg    & \clevr        & \shapestacks   & \cars & \ballsinbowl\\
                           & General       & General       & General       & Ordered        & General  & General\\
\midrule
\blockgantwod &     19.0     &    18.0  &   18.0    & 272.0  & 22.0 & 98.0\\    
Ours                                 & 17.0 & 17.0 & 19.0 & 17.0 & 23.0 & 26.0\\
\bottomrule
\end{tabular}
\end{table}

We have conducted additional experiments to provide more quantitative insights of the disentanglement capabilities of our model.
While measures such as MIG~\citep{locatello2019fairness} are typically used to quantify disentanglement, computing this score is not applicable in our case since our model does not feature an inference component to compute the posterior $q(z|x)$.
Hence, we have devised a proxy procedure: We toggle each object of an image individually (out of 5 objects generated) and measure how the generated image changes.
We report the distance between the pixel location corresponding to the maximum image change and the location (scaled $\theta_i$) of the object that was toggled.
In \cref{t:dis} we report the median distance between $\theta_i$ and the pixel location corresponding to the maximum image change.
We note that our model generally outperforms BlockGAN2D, most notably for \shapestacks where BlockGAN is not able to disentangle different objects at all.
In addition, we note that for the model trained on \shapestacks, the discriminator can predict the position of a stack's base object with 11.3 mean pixel error on the test set.

\subsection{Scale experiment}\label{ssec:scale}

In order to tackle the limitation discussed in \cref{s:experiments}, we propose to augment our model with scale prediction.
Practically, this translates to predicting $H'$ for each individual object instead of keeping it fixed.
Therefore, we now assign $H'_k$ instead of $H'$ to each foreground component of the scene.
$H'_k$ is computed by a module $sc$ following the equation: $H'_k = H' \times (1+sc(z_0, \theta_k, z_k))$.
More details on $sc$ can be found in \cref{tab:sc}.
We summarize all hyper-parameters used for the training of the model in \cref{tab:hyperparams-scale}.
For evaluation, we sample $z_0$ from $\mathcal{U}([-1,1]^{N_b})$, except for the \cars video model where we use $\mathcal{U}([-0.5,0.5]^{N_b})$, a range better suited for optimal background fidelity on this dataset.

\begin{table}[t]
    \centering
    \scriptsize
    \caption{{\bf Network architecture for module $sc$.}}
    \label{tab:sc}
    \begin{tabular}{ccccc}
    \toprule
        {\bf Layer name} & {\bf Layer Type} & {\bf Input size} &{\bf Output size} & {\bf Activation} \\
        \midrule
        FC$sc$\_1 & Linear & $N_f + 2 + N_b $ & 32 & LeakyReLU \\
        \midrule
        FC$sc$\_2 & Linear & 32 & 32 & LeakyReLU \\
        \midrule
        FC$sc$\_3 & Linear & 32 & 1 & Tanh \\
        \bottomrule
    \end{tabular}
\end{table}
\begin{table}[t]
    \centering
    \scriptsize
    \caption{{\bf Hyper-parameters for each datasets for the scale augmented model.} Epoch nums are the number of epochs we trained for. Model is described in \cref{ssec:scale}}
    \label{tab:hyperparams-scale}

    \begin{tabular}{ccccccccccc}
    \toprule
        {\bf Dataset} & {\bf Learning rate} & {\bf Epoch nums } &{\bf $M$} & {$K_{min}-K_{max}$} & { $H'$} & {\bf $N_b$} & {\bf $N_f$} & {\bf $H''/H$ sampling range}\\
        \midrule
        CLEVR3 & 0.0001 & 40 & 2 & 2-3 & 6 & 1 & 20 & $[-0.6,0.6]^2$\\
        \midrule
        CLEVR5 & 0.0001 & 40 & 2 & 2-5 & 4 & 1 & 20 & $[-0.6,0.6]^2$\\
        \midrule
        CLEVR5-vbg & 0.0001 & 40 & 2 & 2-5 & 4 & 1 & 20 & $[-0.6,0.6]^2$\\
        \midrule
        CLEVR & 0.0001 & 40 & 2 & 3-6 & 6 & 1 & 20 & $[-0.6,0.6]^2$\\
        \midrule
        \shapestacks & 0.001 & 30 & 2 & 2-5 & 4 & 5 & 20 & $[-0.6,0.6] \times [0,0.6]$ \\
        \midrule
        \cars & 0.0001 & 20 & 2 & 1-5 & 6 & 1 & 20 & $[-0.6,0.6]^2$\\
        \bottomrule
    \end{tabular}
\end{table}

\begin{figure}[h]
\centering
\includegraphics[width=0.98\textwidth]{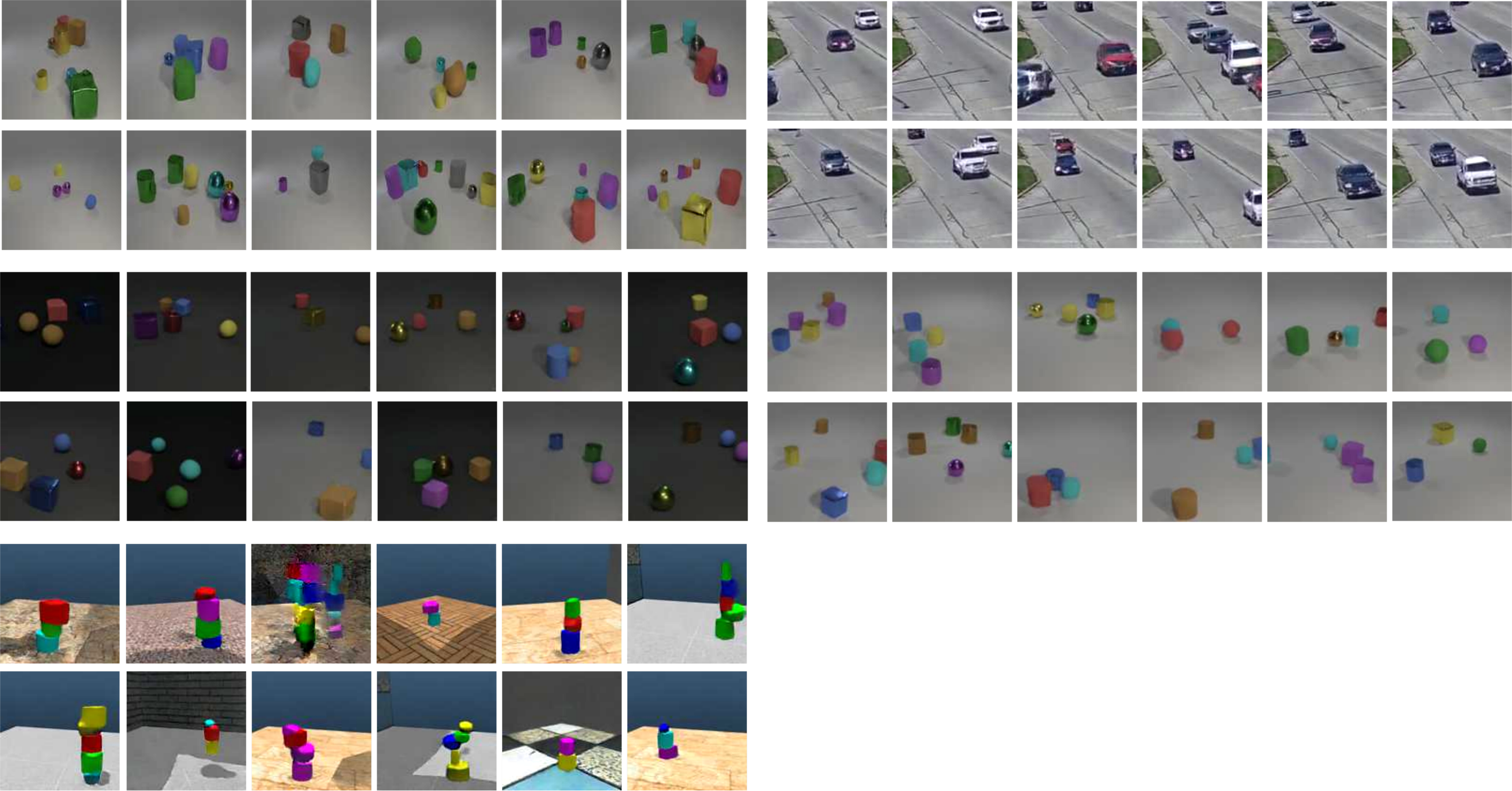}
\caption{\textbf{Samples from the scale augmented RELATE.} From left to right and top to bottom we display CLEVR, \cars, CLEVR-5vbg, CLEVR-5, \shapestacks.  Predictions on CLEVR comprise more variety of object sizes compared to our main model (see \cref{fig:samples_clevr}). Similarly on \cars we can see that cars can be rendered at arbitrary points in the road.}\label{fig:relate_scale}
\end{figure}

\section{Further Discussions}\label{sec:lim_app}
As noted in the conclusions, RELATE is limited to 2D representations.
However, our relaionship module is generic enough to be exported to 3D and could be inserted directly into BlockGAN\cite{nguyen2020blockgan} -- albeit at the expense of significant additional training time.
Finally, we acknowledge that the simplified spatial representation of an object as its centroid position is inferior to other object-centric models which predict full object masks, e.g.~\cite{greff2019iodine,engelcke2019genesis,anciukevicius2020ocf}.
While we believe that an object segmentation mechanism could be easily added based on the already existing rendering of individual latent variables (cf.~\cref{fig:scene-dec}), we conjecture that the simplicity of the centroid representation greatly facilitates the learning of spatial correlations within $\Gamma$.
In contrast, using object instance masks as spatial representations might introduce unnecessary complications for the neural physics approximation.

\section{Losses}\label{sec:losses}
Our final loss is the sum of three losses mentioned in the main text:
$$
\mathcal{L}_{tot} =  \mathcal{L}_{\text{GAN}}(\hat I, I) + \mathcal{L}_{\text{style}}(\hat I, I) + \min_{G,\Gamma,P}\| \check{\theta}_k - P(G(W(z_0, z_k, \theta_k))) \|^2_2,
$$
where $ \mathcal{L}_{\text{GAN}}(\hat I, I)$ is the standard GAN loss:
$$
 \mathcal{L}_{\text{GAN}}(\hat I, I) = \min_{G,\Gamma} \max_{D} \mathbb{E}[ \log(1-D(G(W(Z, \Theta)))] + \mathbb{E}[\log(D(I))]
$$
The style loss follows the implementation of BlockGAN~\cite{nguyen2020blockgan}.
The input of the style discriminator $D_l$ are mean $\mu_l$ and variance $\sigma_l^2$ across spatial dimensions of $\Phi_l(x) \in \mathcal{R}^{W_l\times H_l \times C_l}$, the output of the $l$th layer of $D$  taken before the normalization step:
$$
\mu_l(\Phi_l (x)) = \frac{1}{W_l \times H_l}\sum_i\sum_j  \Phi_l(x)_{i,j} ,
$$
$$
\sigma_l^2(\Phi_l (x)) = \frac{1}{W_l \times H_l}\sum_i\sum_j  (\Phi_l(x)_{i,j} - \mu_l(\Phi_l (x)) )^2.
$$
The style discriminator $D_l$ for each layer is then implemented as a linear layer followed by a sigmoid activation function.
The resulting style loss is:
$$
\mathcal{L}_{\text{style}}(\hat I, I) = \max_{D} \sum_l \mathbb{E}[ \log(1-D_l(\hat I))] + \mathbb{E}[\log(D_l(I))]
$$

\section{Implementation details}\label{sec:app:impldet}
\paragraph{Infrastructure and framework} For all experiments we use PyTorch 1.4. We train all models on a single NVIDIA Tesla V100 GPU.

\paragraph{Training hyperparameters} We initialize all weights (including instance normalization ones) by drawing from a random normal distribution $\mathcal{N}(0,0.02)$. All biases were initialized to 0.
For each update of the discriminator we update the generator $M$ number of times. We use Adam parameters $(\beta_1, \beta_2) = (0., 0.999)$ for all datasets except \ballsinbowl where $\beta_1 = 0.5$. Similarly $W$ was a max-pooling operator in all datasets except \ballsinbowl where we used a sum pooling operator.
As in BlockGAN\cite{nguyen2020blockgan}, background and foreground decoders each start from a learned constant tensors $T_b$ and $T_f$ respectively with sizes $H\times H\times 256$ and $H' \times H' \times 512$. For \ballsinbowl we use a tensor $T_f$ for each object and use a constant style vector of one.

In the case of dynamic scenarios we reuse same hyperparameters as in the static case except that we use a learning rate of 0.0001 and $\beta_1=0.$

Full details of the parameters for each dataset can be found in \cref{tab:hyperparams}

\paragraph{Evaluation details}
For FID scores computation we draw 10\,000 samples from our model which we compare against the same number of images drawn from the test set.
To compute FVD score on each dataset, we sample 500 videos of 30 frames from our model and compare them against the videos of the respective test sets (500 for \ballsinbowl and 275 videos for \cars). This also applies to the time shuffled baseline.

To be able to compare with other methods we resize our generated images to $96 \times 96$ on CLEVR5 and CLEVR5-vbg and $64 \times 64$ for \shapestacks.
For the simple generative baselines, DRAGAN and DCGAN we evaluate FID score on the generated $64\times64$ images from these models.
We evaluate on the generated $128 \times 128$ images otherwise.

We empirically found that background was rendered with better quality for lower values of $z_0$. Hence at test time we sampled $z_0$ from $\mathcal{U}([-0.5,0.5]^{N_b})$ for optimal results.

\begin{table}[t]
    \centering
    \scriptsize
    \caption{{\bf Hyper-parameters for each datasets.} Epoch nums are the number of epochs we trained for. }
    \label{tab:hyperparams}

    \begin{tabular}{ccccccccccc}
    \toprule
        {\bf Dataset} & {\bf Learning rate} & {\bf Epoch nums } &{\bf $M$} & {$K_{min}-K_{max}$} & { $H'$} & {\bf $N_b$} & {\bf $N_f$} & {\bf $H''/H$ sampling range}\\
        \midrule
        \ballsinbowl & 0.001 & 60 & 1 & 2-2 & 8 & 3 & 1 & $[-0.8,0.8]^2$ \\
        \midrule
        CLEVR5 & 0.0001 & 40 & 2 & 2-5 & 4 & 1 & 90 & $[-0.6,0.6]^2$\\
        \midrule
        CLEVR5-vbg & 0.0001 & 30 & 2 & 2-5 & 4 & 1 & 90 & $[-0.6,0.6]^2$\\
        \midrule
        CLEVR & 0.0001 & 40 & 2 & 3-6 & 4 & 1 & 90 & $[-0.6,0.6]^2$\\
        \midrule
        \shapestacks & 0.001 & 30 & 2 & 2-5 & 4 & 12 & 64 & $[-0.6,0.6] \times [0,0.6]$ \\
        \midrule
        \cars & 0.0001 & 20 & 2 & 1-5 & 6 & 1 & 20 & $[-0.6,0.6]^2$\\
        \bottomrule
    \end{tabular}
\end{table}


\subsection{Architecture details}\label{ss:arch}
\paragraph{Generator.} In this work we maintain the core of our architecture fixed as much as possible. Since the dimension of the sample $z_i$ does not necessarily match the channel dimension where it is injected before applying Adaptive Instance Normalisation (AdaIN) to a layer $l$ we map $z_i$ to a vector $\hat z_i$ transformed such that
$$
\hat z_i = \max(W_l^T z_i + b_l, 0)
$$
Where $(W_l,b_l)$ are learnable parameters. AdaIN is applied at the end of the layers (after the activation).
All LeakyReLU layers are using a parameter of 0.2.

\begin{table}[t]
    \centering
    \scriptsize
    \caption{{\bf Network architecture for the foreground object generator $\Psi_f$.}}
    \label{tab:psif}
    \begin{tabular}{cccccccc}
    \toprule
        {\bf Layer name} & {\bf Layer Type} & {\bf Input size} &{\bf Output size} & {\bf Kernel Size} & {\bf Stride} & {\bf Activation} & {\bf Norm.} \\
        \midrule
        Style\_f & Id & $H'\times H'\times 512$ & $H'\times H'\times 512$ & - & - & Id & AdaIn \\
        \midrule
        Convtf\_1 &  ConvTranspose & $H'\times H'\times 512$  & $H'\times H'\times 512$  & $3 \times 3$ & 1 & LeakyReLU & AdaIn \\
        \midrule
        Convtf\_2 &  ConvTranspose & $H'\times H'\times 512$  & $H'\times H'\times 256$  & $3 \times 3$ & 1 & LeakyReLU & AdaIn \\
        \midrule
        Pad & Padding & $H'\times H'\times 512$  & $H\times H\times 256$  & - & - & - & -\\
        \bottomrule
    \end{tabular}
\end{table}
\begin{table}[t]
    \centering
    \scriptsize
    \caption{{\bf Network architecture for the background object generator $\Psi_b$.}}
    \label{tab:psib}

    \begin{tabular}{cccccccc}
    \toprule
        {\bf Layer name} & {\bf Layer Type} & {\bf Input size} &{\bf Output size} & {\bf Kernel Size} & {\bf Stride} & {\bf Activation} & {\bf Norm.} \\
        \midrule
        Style\_b & Id & $H\times H\times 256$ & $H\times H\times 512$ & - & - & Id & AdaIn \\
        \midrule
        Convtb\_1 &  ConvTranspose & $H\times H\times 512$  & $H\times H\times 512$  & $3 \times 3$ & 1 & LeakyReLU & AdaIn \\
        \midrule
        Convtb\_2 &  ConvTranspose & $H\times H\times 512$  & $H\times H\times 256$  & $3 \times 3$ & 1 & LeakyReLU & AdaIn \\
        \bottomrule
    \end{tabular}
\end{table}

\begin{table}[t]
    \centering
    \scriptsize
    \caption{{\bf Network architecture for the generator G.} Outputs of all $K$ foreground object generators $\Psi_f$ and background generator $\Psi_b$ are stacked on the first dimension before entering layer $W$ (third row).}
    \label{tab:generator}
    \begin{tabular}{ccccccc}
    \toprule
        {\bf Layer name} & {\bf Layer Type} & {\bf Input size} &{\bf Output size} & {\bf Kernel Size} & {\bf Stride} & {\bf Activation}\\
        \midrule
        \tikzmark{psif}$\Psi_f$ (see \cref{tab:psif})  & - & $H'\times H'\times 512$ & $16\times 16\times 256$ & - & - & -  \\
        \midrule
        \tikzmark{psib}$\Psi_b$ (see \cref{tab:psib}) & - & $H\times H\times 512$ & $16\times 16\times 256$ & - & - & -  \\
        \midrule
        \tikzmark{w}W & Max/Sum Pool & $(K+1) \times 16\times 16\times 256$ & $16\times 16\times 256$ & - & - & -  \\
        \midrule
        Convtg\_1 &  ConvTranspose & $16\times 16\times 256$  & $32\times 32\times 128$  & $4 \times 4$ & 2 & LeakyReLU \\
        \midrule
        Convtg\_2 &  ConvTranspose & $32\times 32\times 128$  & $64\times 64\times 64$  & $4 \times 4$ & 2 & LeakyReLU \\
        \midrule
        Convtg\_3 &  ConvTranspose & $64\times 64\times 64$  & $64\times 64\times 64$  & $3 \times 3$ & 1 & LeakyReLU \\
        \midrule
        Convtg\_4 &  ConvTranspose & $64\times 64\times 64$  & $128\times 128\times 64$  & $4 \times 4$ & 2 & LeakyReLU \\
        \midrule
        Convtg\_5 &  ConvTranspose & $128\times 128\times 64$  & $128\times 128\times 3$  & $3 \times 3$ & 1 & Tanh \\
        \bottomrule
    \end{tabular}
    \begin{tikzpicture}[overlay, remember picture, shorten >=.25pt, shorten <=.25pt]
    \draw [->] ([yshift=.25\baselineskip]{pic cs:psif}) [bend right=80] to ([xshift=-3em,yshift=.25\baselineskip]{pic cs:w});
    \draw [->] ([yshift=.25\baselineskip]{pic cs:psib}) [bend right=60] to ([xshift=-2.6em,yshift=.25\baselineskip]{pic cs:w});
  \end{tikzpicture}
\end{table}

\begin{table}[t]
    \centering
    \scriptsize
    \caption{{\bf Network architecture for module $f$ and $f_v$.} \textsuperscript{*} indicates modification of $f_v$}
    \label{tab:f}
    \begin{tabular}{ccccc}
    \toprule
        {\bf Layer name} & {\bf Layer Type} & {\bf Input size} &{\bf Output size} & {\bf Activation} \\
        \midrule
        FC$f$\_1 & Linear & $2\times(N_f + 2 + 2^{*})$ & 32 & LeakyReLU \\
        \midrule
        FC$f$\_2 & Linear & 32 & 32 & LeakyReLU \\
        \midrule
        FC$f$\_3 & Linear & 32 & 32 & None \\
        \bottomrule
    \end{tabular}
\end{table}

\begin{table}[t]
    \centering
    \scriptsize
    \caption{{\bf Network architecture for module $g$ and $g_v$.} \textsuperscript{*} indicates modification of $g_v$}
    \label{tab:g}
    \begin{tabular}{ccccc}
    \toprule
        {\bf Layer name} & {\bf Layer Type} & {\bf Input size} &{\bf Output size} & {\bf Activation} \\
        \midrule
        FC$g$\_1 & Linear & $32+ N_f + 2 +  2^{*} + N_b $ & 32 & LeakyReLU \\
        \midrule
        FC$g$\_2 & Linear & 32 & 32 & LeakyReLU \\
        \midrule
        FC$g$\_3 & Linear & 32 & 2 & Tanh \\
        \bottomrule
    \end{tabular}
\end{table}

\begin{table}[t]
    \centering
    \scriptsize
    \caption{{\bf Network architecture for module $e_v$.}}
    \label{tab:e}
    \begin{tabular}{ccccc}
    \toprule
        {\bf Layer name} & {\bf Layer Type} & {\bf Input size} &{\bf Output size} & {\bf Activation} \\
        \midrule
        FC$e_v$\_1 & Linear & $N_f + 2 + N_b $ & 128 & LeakyReLU \\
        \midrule
        FC$e_v$\_2 & Linear & 128 & 128 & LeakyReLU \\
        \midrule
        FC$e_v$\_3 & Linear & 128 & $3 \times 2$ & Tanh \\
        \bottomrule
    \end{tabular}
\end{table}

\begin{minipage}{0.45\linewidth}
\centering
    \centering
    \scriptsize
    \captionof{table}{{\bf Network architecture for module $f_0$.}}
    \label{tab:f0}
    \begin{tabular}{p{0.6cm}p{0.5cm}p{1.4cm}p{0.7cm}p{1.1cm}}
    \toprule
        {\bf Layer name} & {\bf Layer Type} & {\bf Input size} &{\bf Output size} & {\bf Activation} \\
        \midrule
        FC$f_0$\_1 & Linear & $N_f + N_b +2$ & 128 & LeakyReLU \\
        \midrule
        FC$f_0$\_2 & Linear & 128 & 64 & LeakyReLU \\
        \midrule
        FC$f_0$\_3 & Linear & 64 & 2 & Tanh \\
        \bottomrule
    \end{tabular}
\end{minipage}
\begin{minipage}{0.55\linewidth}
\centering
    \centering
    \scriptsize
    \captionof{table}{{\bf Network architecture for module $f_1$.}}
    \label{tab:f1}
    \begin{tabular}{p{0.8cm}p{1.2cm}p{1.4cm}p{0.5cm}p{1.1cm}}
    \toprule
        {\bf Layer name} & {\bf Layer Type} & {\bf Input size} &{\bf Output size} & {\bf Activation} \\
        \midrule
        FC$f_1$\_1 & Linear & $N_f + N_b$ & 128 & LeakyReLU \\
        \midrule
        FC$f_1$\_2 & Linear & 128 & 64 & LeakyReLU \\
        \midrule
        FC$f_1$\_3 & Linear & 64 & 2 & None \\
        \midrule
        $Pos_{out}$ &  Sigmoid(x) & 2 & 2 & None \\
        & Tanh(y) \\
        \bottomrule
    \end{tabular}
\end{minipage}

\paragraph{Discriminator}
We describe the architecture of the discriminator network in more details in \cref{tab:decoder}. We use spectral normalization~\cite{miyato2018spectral} at almost every layer. Positions are directly regressed from the last feature output of the discriminator (see last line $P_{end}$). Therefore in practice $P$ and $D$ share the same backbone $D_b$ (see table \cref{tab:decoder} until flatten) for every image I:
$$
P(I) = P_{end}(D_b(I)), ~~~~~~ D(I) = Disc(D_b(I)).
$$

Input for style discriminator are taken after the convolution of (Convd\_2, Convd\_3, Convd\_4, Convd\_5) in \cref{tab:decoder} before the normalization. Spectral Normalization was \emph{not} applied to any $D_l$.

\begin{table}[ht]
    \centering
    \scriptsize
    \caption{{\bf Network architecture for the discriminators.} Note that the Instance Normalization weights were also subjected to spectral normalization. $P$ and $D$ shares weights until Flatten layer.}
    \label{tab:decoder}

    \begin{tabular}{cccccccc}
    \toprule
        {\bf Layer name} & {\bf Layer Type} & {\bf Input size} &{\bf Output size} & {\bf Kernel Size} & {\bf Stride} & {\bf Activation} & {\bf Norm.} \\
        \midrule
        Convd\_1 & Conv & $128\times 128\times 3$ & $64\times 64\times 64$ & $5 \times 5$ & 2 & LeakyReLU & -  \\
        \midrule
        Convd\_2 & Conv & $64\times 64\times 64$ & $32\times 32\times 128$ & $5 \times 5$ & 2 & LeakyReLU & IN/Spec. Norm.  \\
        \midrule
        Convd\_3 & Conv & $32\times 32\times 128$ & $16\times 16\times 256$ & $5 \times 5$ & 2 & LeakyReLU & IN/Spec. Norm.  \\
        \midrule
        Convd\_4 & Conv & $16\times 16\times 256$ & $8\times 8\times 512$ & $5 \times 5$ & 2 & LeakyReLU & IN/Spec. Norm.  \\
        \midrule
        Convd\_5 & Conv & $8\times 8\times 512$ & $4\times 4\times 1024$ & $5 \times 5$ & 2 & LeakyReLU & IN/Spec. Norm.  \\
        \midrule
        \tikzmark{flatten}Flatten & Id & $4\times 4\times 1024$ & $1\times 1\times 16384$ & - & - & - & -  \\
        \midrule
        \midrule
        \tikzmark{disc} $Disc$ & Linear & $1\times 1\times 16384$ & $1$ & - & - & Sigmoid & None/Spec. Norm.  \\
        \midrule
        \tikzmark{pend} $P_{end}$ & Linear & $1\times 1\times 16384$ & $2$ & - & - & Tanh & None/Spec. Norm.  \\
        \bottomrule
    \end{tabular}
    \begin{tikzpicture}[overlay, remember picture, shorten >=.25pt, shorten <=.25pt]
    \draw [->] ([yshift=.25\baselineskip]{pic cs:flatten}) [bend right=50] to ([yshift=.25\baselineskip]{pic cs:disc});
    \draw [->] ([yshift=.25\baselineskip]{pic cs:flatten}) [bend right=80] to ([yshift=.25\baselineskip]{pic cs:pend});
  \end{tikzpicture}
\end{table}

\section{Baselines}\label{sec:baselines}
\paragraph{DCGAN\citep{radford2015unsupervised} and DRAGAN\citep{kodali2017convergence}.} We used an online pytorch implementation\footnote{\url{https://github.com/LynnHo/DCGAN-LSGAN-WGAN-GP-DRAGAN-Pytorch}} with default hyper-parameters.
We trained these models to generate $64\times64$ images and therefore only evaluated FID score at the same resolution (see \cref{sec:app:impldet}).

\paragraph{OCF.} OCF results were copied from original paper of \cite{anciukevicius2020ocf}.

\paragraph{BlockGAN2D.} We use the same hyperparameters and network architecture as RELATE except for learning rate and $M$. In all cases we report the best results over models trained with variations of learning rate in (0.001, 0.0001) and $M$ in (2,3).

\paragraph{GENESIS.}
We use the official implementation\footnote{\url{https://github.com/applied-ai-lab/genesis}} of \genesis for all experiments.
For the ShapeStacks dataset, we use the official model snapshot released with the original paper\footnote{\url{https://drive.google.com/drive/folders/1uLSV5eV6Iv4BYIyh0R9DUGJT2W6QPDkb?usp=sharing}}.
For all other datasets, we train \genesis for 500,000 iterations with the default learning parameters and select the last model checkpoint for evaluation.
When training \genesis we use \emph{constrained ELBO optimization}~\cite{rezende2018geco} controlled via \texttt{g\_goal} in the training script which influences the decomposition capability of \genesis.
We perform a grid search over \texttt{g\_goal} in the range of 0.5635 to 0.5655 and select the model with the lowest ELBO after 500,000 iterations.

\section{Datasets}\label{sec:dataset}

\paragraph{\ballsinbowl.}
\begin{figure}[t]
\centering
\includegraphics[width=0.7\textwidth]{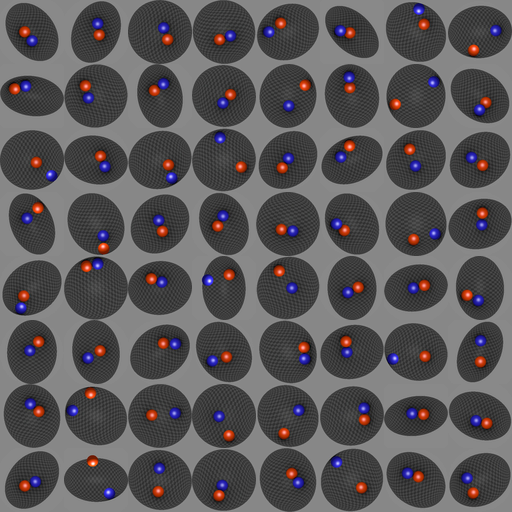}
\caption{\textbf{Sample data from \ballsinbowl.} The dataset consists of two balls of different colors rolling in elliptical bowls of various shapes.}\label{fig:bowls}
\end{figure}
This dataset is a replica of the two balls synthetic dataset of \cite{ehrhardt2019taking}. It consists of 2500 training sequences and 500 test sequences of two balls of different fixed colour rolling in bowls of various shapes. We count an epoch as 10,000 iterations over the data. In \cref{fig:bowls} we show some sample data from this dataset.

\paragraph{CLEVR.} We used the official CLEVR from \cite{johnson2017clevr}. We train on data from train and validation set and evaluate on the test set. Both ours and BlockGAN2D were trained on the subset containing 3 to 6 objects and evaluated on the entire test set.

\paragraph{CLEVR5/CLEVR5-vbg.} We use online code provided by the authors\footnote{\url{https://github.com/TitasAnciukevicius/clevr-dataset-gen.}} to generate CLEVR5 and CLEVR5-vbg. As done in \cite{anciukevicius2020ocf} we generate 100,000 images keep 90,000 for training and 10,000 for testing.

\paragraph{ShapeStacks}
We use the official release of the \shapestacks dataset\footnote{\url{https://shapestacks.robots.ox.ac.uk/\#data}}.
We use the \texttt{default} partitioning provided with the dataset and merge the training and validation splits for a total of 264,384 training images.
All FID comparisons are made against 10,000 images randomly sampled from the test set which contains 46,560 images in total.
Since the original resolution of the images is $224 \times 224$ pixels, we re-scale them to $128 \times 128$ before feeding them to our network.

\paragraph{\cars.} We recorded 5 hours from Youtube\footnote{\url{https://www.youtube.com/watch?v=5_XSYlAfJZM}} of a live traffic camera at a crossing. The video was then unrolled at 10 fps and manually processed to keep only sequences with a number of cars in [1,5]. We kept 560 videos for the training set and 123 in test (80/20 ratio). This dataset will be publicly released.


\section{Qualitative results}\label{sec:qr}
\begin{figure}[t]
\centering
\includegraphics[width=\textwidth]{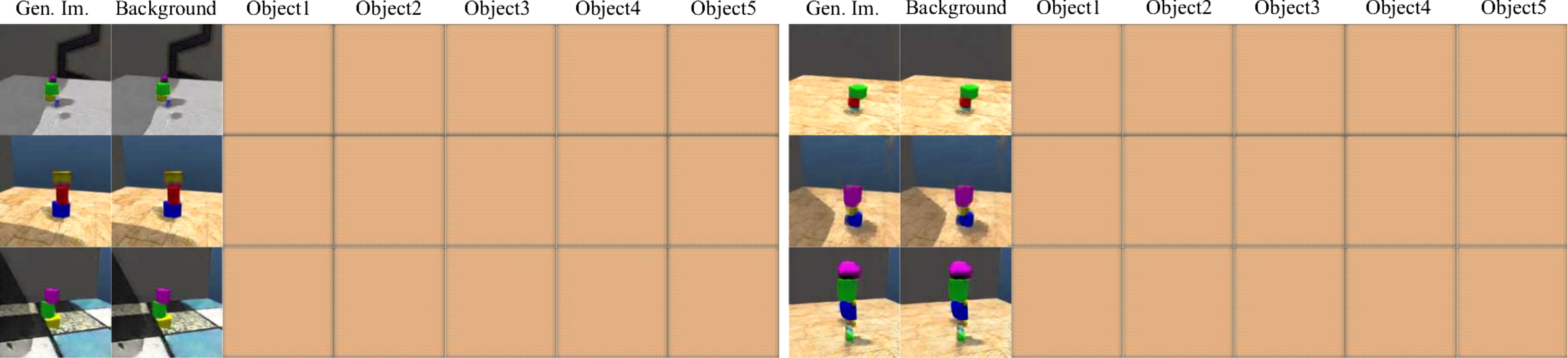}
\caption{\textbf{BlockGAN2D scene decomposition on \shapestacks.} We display in order (generated, background, object\_1, ..., object\_5) for BlockGAN2D model. We see that in this case BlockGAN doesn't capture objectness at all and render everything in the background. This shows how, for structured scenes, prior work fails to capture correlations between objects.}\label{fig:blckgan_shst}
\end{figure}

\begin{figure}[t]
\centering
\includegraphics[height=0.7\textheight]{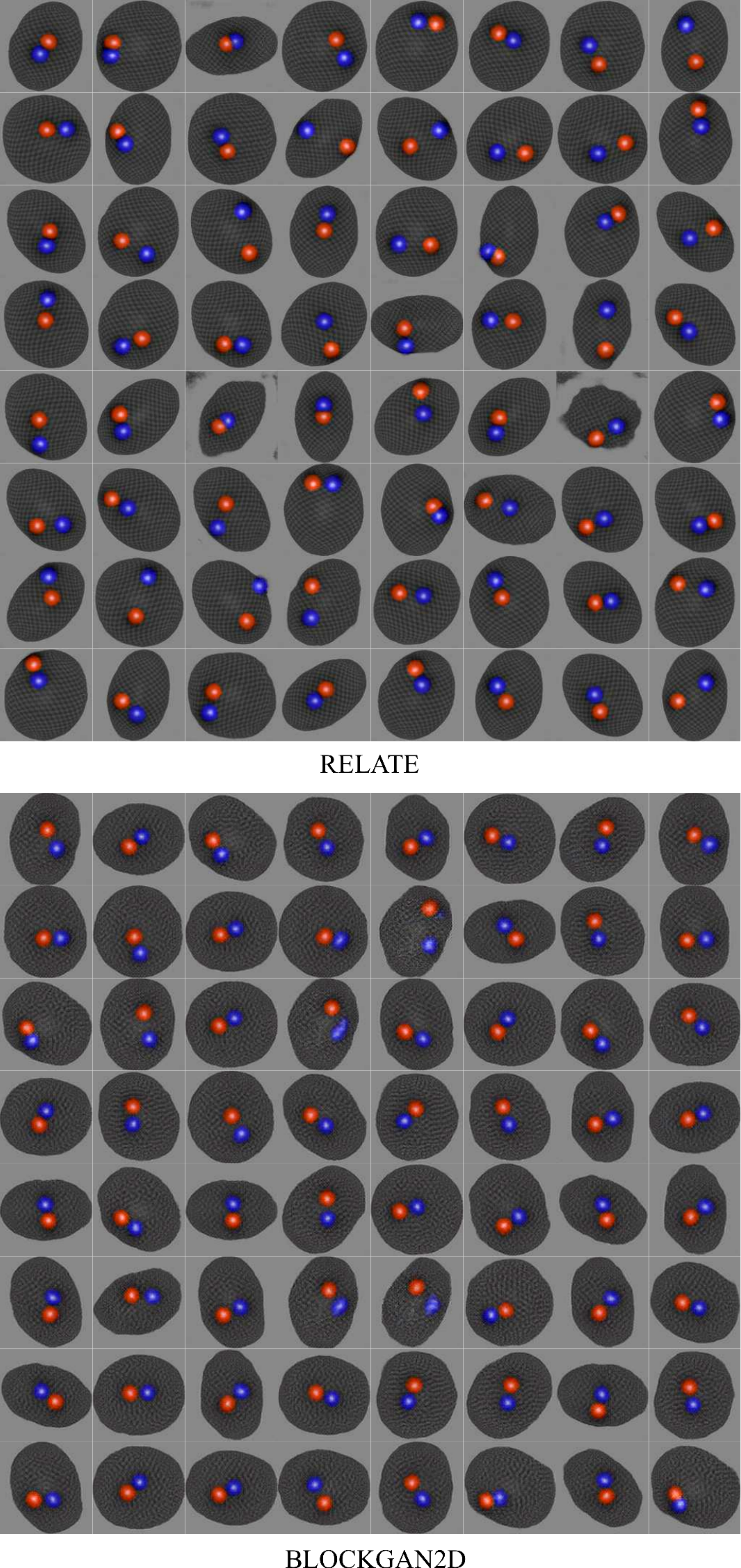}
\caption{\textbf{Generated scenes for models trained on \ballsinbowl.} Qualitatively RELATE generates images of higher quality compared to BlockGAN2D\cite{nguyen2020blockgan}.}\label{fig:samples_balls}
\end{figure}

\begin{figure}[t]
\centering
\includegraphics[height=0.98\textheight]{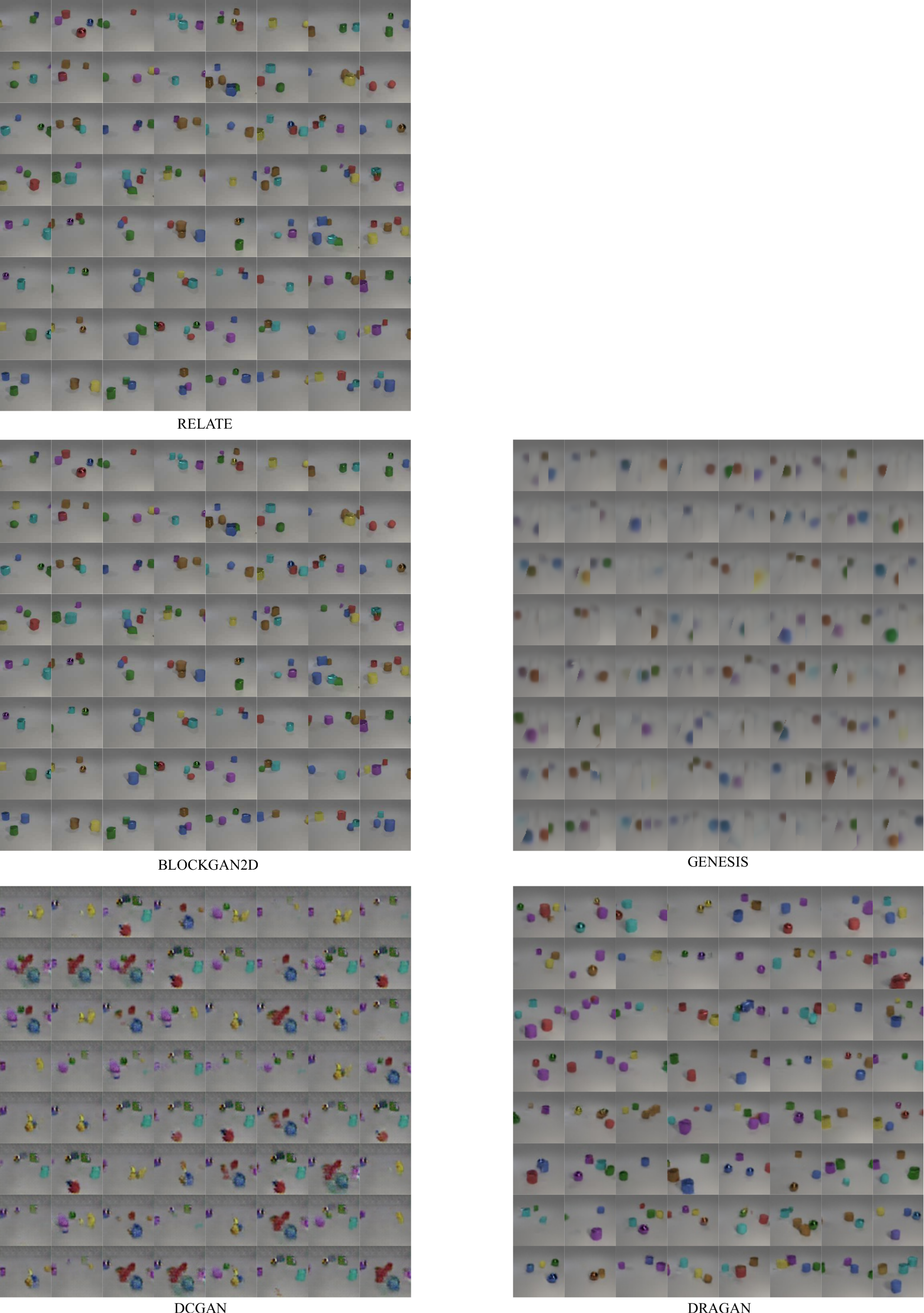}
\caption{\textbf{Generated scenes for models trained on CLEVR5.} For less crowded scenes our model and BlockGAN2D reach similar performances.}\label{fig:samples_clevr5}
\end{figure}

\begin{figure}[t]
\centering
\includegraphics[height=0.98\textheight]{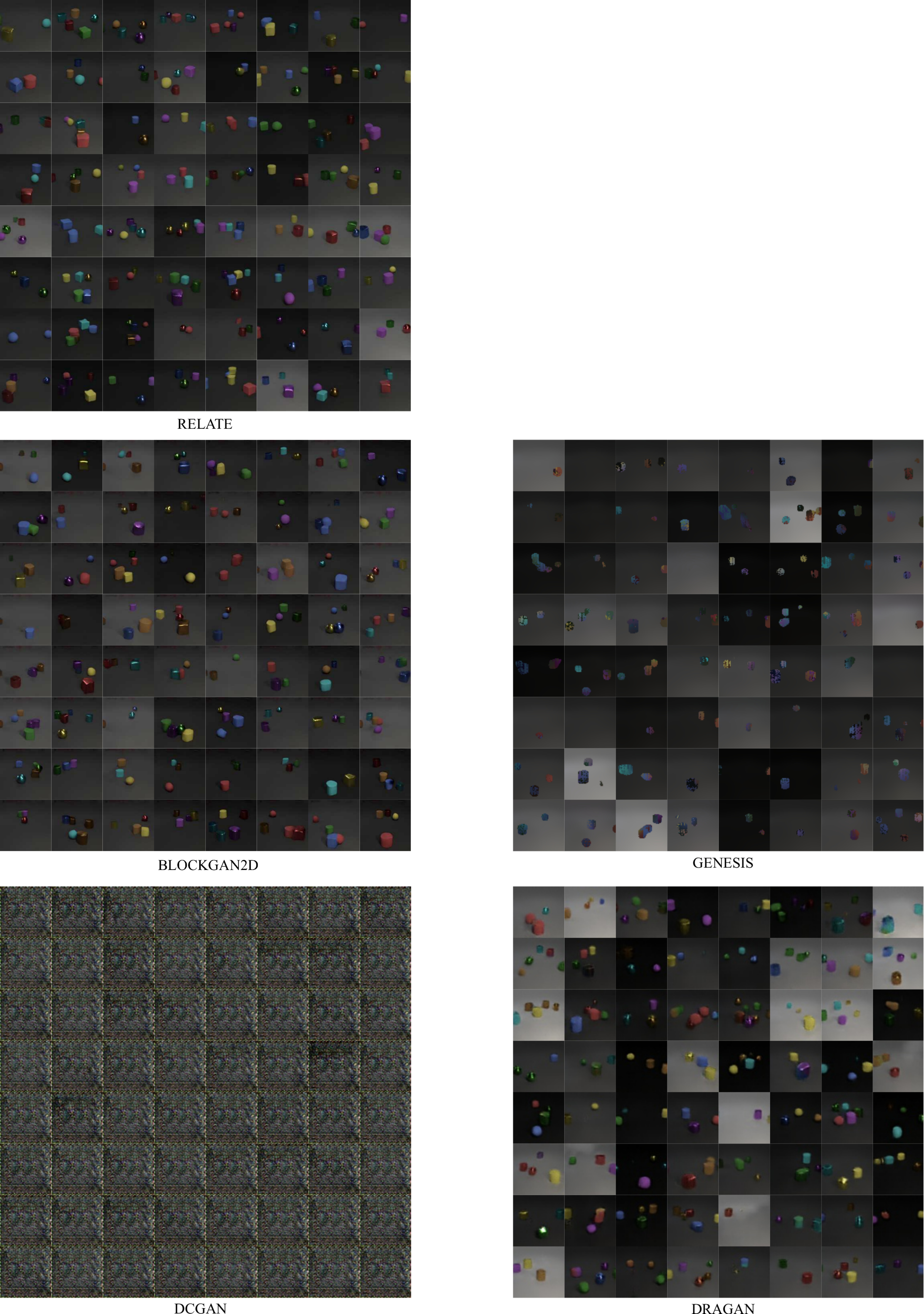}
\caption{\textbf{Generated scenes for models trained on CLEVR5-vbg.} This scenario reaches similar conclusion as \cref{fig:samples_clevr5}.}\label{fig:samples_clevr5vbg}
\end{figure}

\begin{figure}[t]
\centering
\includegraphics[height=0.98\textheight]{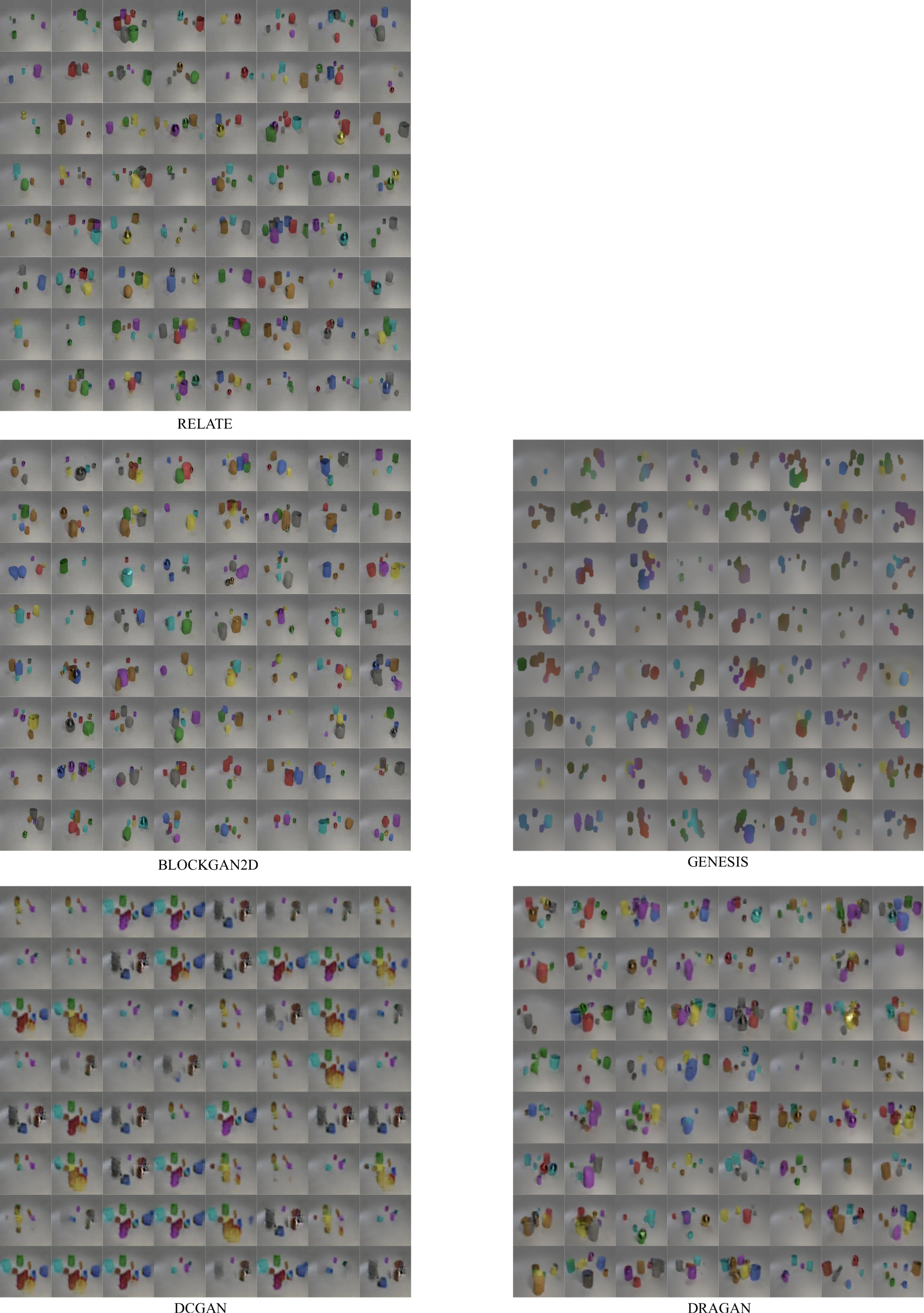}
\caption{\textbf{Generated scenes for models trained on CLEVR.} When the scene gets more crowded RELATE gets an advantage as it can push objects apart resulting in higher qualitative rendering.}\label{fig:samples_clevr}
\end{figure}

\begin{figure}[t]
\centering
\includegraphics[height=0.98\textheight]{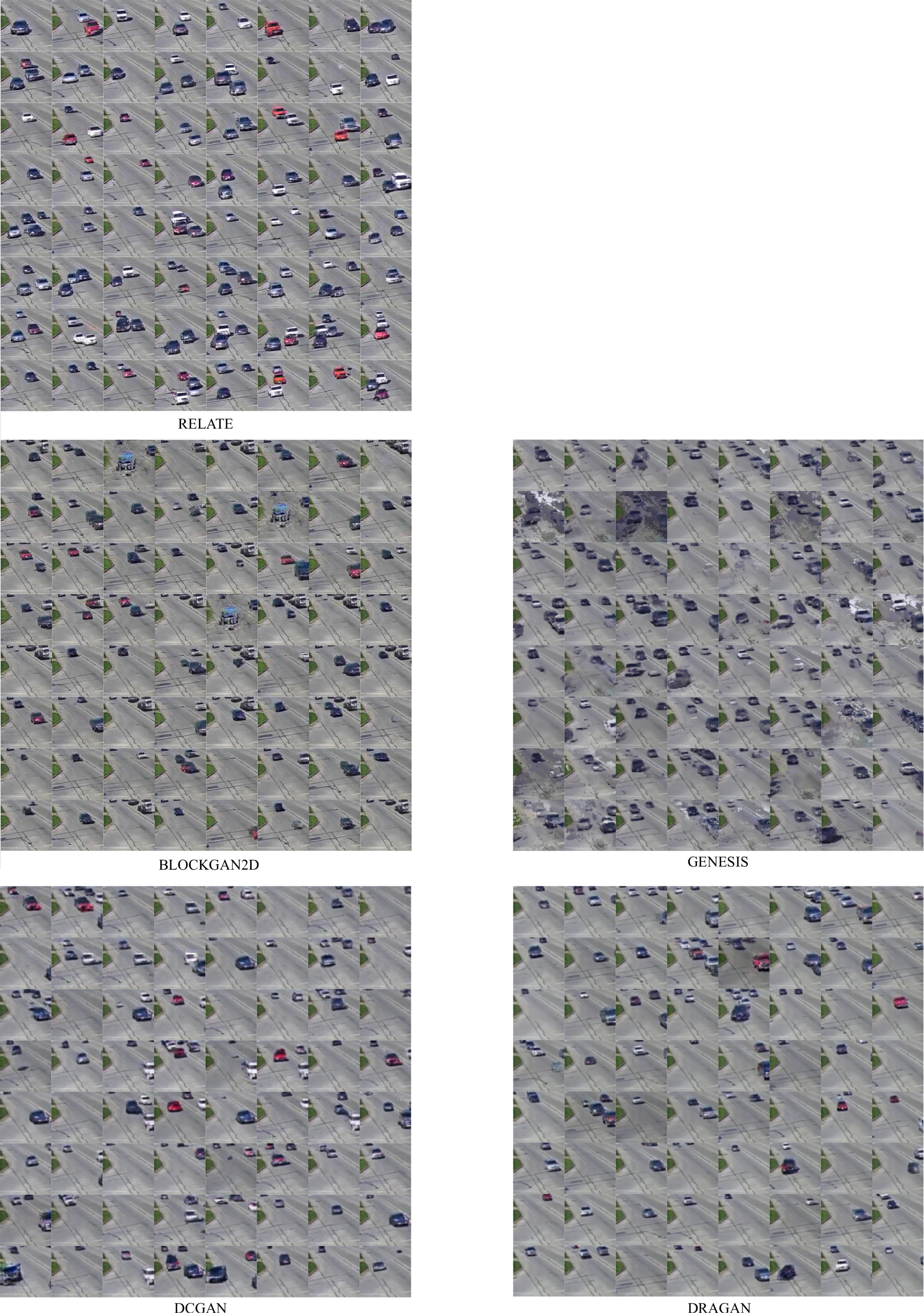}
\caption{\textbf{Generated scenes for models trained on \cars.} Our model qualitatively renders higher fidelity images. BlockGAN2D sometimes suffers from background mode collapse (see first, second and fourth rows of second block).}\label{fig:samples_cars}
\end{figure}

We provide additional qualitative generation results. \Cref{fig:blckgan_shst} shows a failure case of BlockGAN2D mentionned in the paper. In fact, when the scene is more structured BlockGAN2D fails to be object centric and let the background render the entire scene. In addition \cref{fig:samples_balls,fig:samples_clevr,fig:samples_clevr5,fig:samples_clevr5vbg,,fig:samples_cars} provide more samples on every dataset for all the models we trained. In particular we can see that when inter-objects relations are weak in CLEVR5 or CLEVR5-vbg, BlockGAN2D performs qualitatively similar to ours (see \cref{fig:samples_clevr5,,fig:samples_clevr5vbg}). However when the scene is more crowded and the objects have higher correlation BlockGAN2D quality decreseases significantly (see \cref{fig:samples_balls,fig:samples_clevr,,fig:samples_cars}).

\begin{figure}[t]
\centering
\includegraphics[height=0.98\textheight]{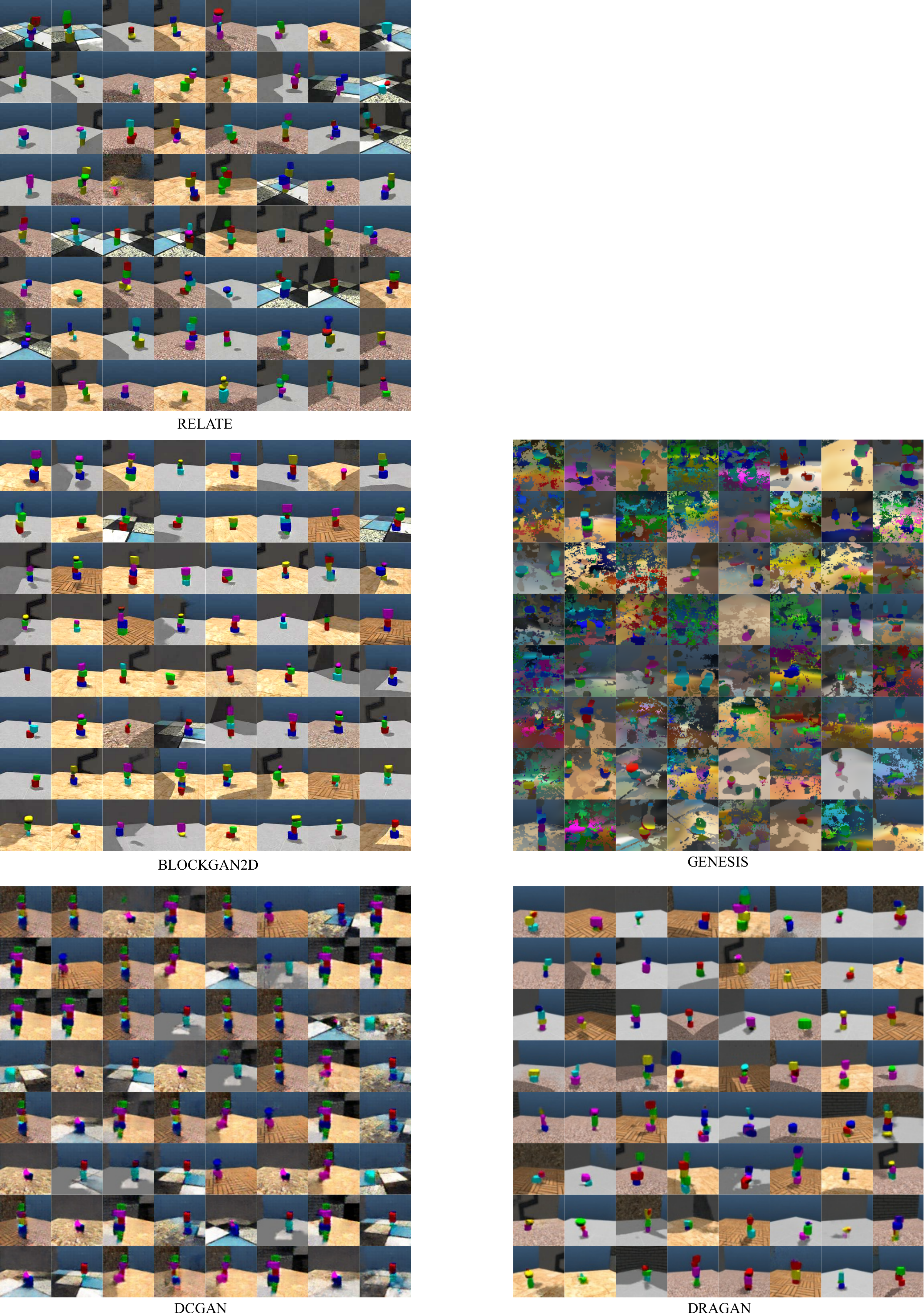}
\caption{\textbf{Generated scenes for models trained on \shapestacks.} Despite qualitative similar rendering, BlockGAN2D isn't rendering a scene component-wise as opposed to ours (see \cref{fig:blckgan_shst}).}\label{fig:samples_shapestacks}
\end{figure}

\end{document}